\documentclass[10pt,journal,compsoc]{IEEEtran}

\usepackage{graphicx}
\usepackage{amsmath}
\usepackage{amssymb}
\usepackage{mathtools}
\usepackage{amsthm}
\usepackage{bm}
\usepackage{multirow}
\usepackage{stfloats}
\usepackage{xcolor}
\usepackage{booktabs}
\usepackage{algorithm}
\usepackage{algorithmic}
\usepackage{cite}
\usepackage[hidelinks]{hyperref}
\usepackage[capitalize,noabbrev]{cleveref}

\theoremstyle{plain}

\theoremstyle{definition}

\theoremstyle{remark}

\makeatletter
\newcommand{\thickhline}{%
    \noalign {\ifnum 0=`}\fi \hrule height 1.2pt
    \futurelet \reserved@a \@xhline
}
\makeatother
\title{MSD-Score: Multi-Scale Distributional Scoring for Reference-Free Image Caption Evaluation}

\author{Shichao Kan, Xuyang Zhang, Haojie Zhang, Zhe Zhu, Yigang Cen, Yixiong Liang,\\ Lianlei Shan, Linna Zhang, Zhe Qu, Jiazhi Xia

\thanks{This work was supported in part by the National Natural Science Foundation of China under Grant 62473033 and 62463002, in part by the Beijing Natural Science Foundation under grant L231012. We are grateful to the High Performance Computing Center of Central South University for partial support of this work.}
\thanks{Corresponding author: Jiazhi Xia}
\thanks{Code is available at: \url{https://steinsgatesg.github.io/MSDScore/}.}
\thanks{Shichao Kan, Xuyang Zhang, Haojie Zhang, Zhe Zhu, Yixiong Liang, Zhe Qu and Jiazhi Xia are with the School of Computer Science and Engineering, Central South University, Changsha, 410083, China (E-mail: kanshichao@csu.edu.cn, 8210231905@csu.edu.cn, 234711081@csu.edu.cn, zhuzhe@csu.edu.cn, yxliang@csu.edu.cn, zhe\_qu@csu.edu.cn, xiajiazhi@csu.edu.cn).}
\thanks{Yigang Cen is with the School of Computer
Science and Technology and the State Key Laboratory of Advanced Rail
Autonomous Operation, Beijing Jiaotong University, Beijing, 100044, China
(E-mail: ygcen@bjtu.edu.cn).}
\thanks{Lianlei Shan is with the School of Computer Science and Technology, University of Chinese Academy of Sciences, Beijing, 101408, China (E-mail: shanlianlei18@mails.ucas.edu.cn)}
\thanks{Linna Zhang is with the School of Mechanical Engineering, Guizhou University, Guiyang, 550025, China (E-mail: zln770808@163.com)}
}

\begin{document}

\maketitle

\begin{abstract}
Evaluating image captions without references remains challenging because global embedding similarity often misses fine-grained mismatches such as hallucinated objects, missing attributes, or incorrect relations. We propose \textbf{MSD-Score}, a reference-free metric that models image patch and text token embeddings as von Mises--Fisher mixtures on the unit hypersphere. Instead of treating each modality as a single point, MSD-Score formulates image--text matching as a \emph{multi-scale distributional scoring} problem. Semantic discrepancies are quantified via a weighted bi-directional KL divergence and combined with global similarity in a multi-scale framework for both single- and multi-candidate evaluations. Extensive experiments show that MSD-Score achieves state-of-the-art correlation with human judgments among reference-free metrics. Beyond accuracy, its probabilistic formulation yields transparent and decomposable diagnostics of local grounding errors, providing a deterministic complementary signal to holistic similarity metrics and judge-based evaluators.
\end{abstract}


\section{Introduction}

\begin{figure}[!t]
    \centering
    \includegraphics[width=\columnwidth]{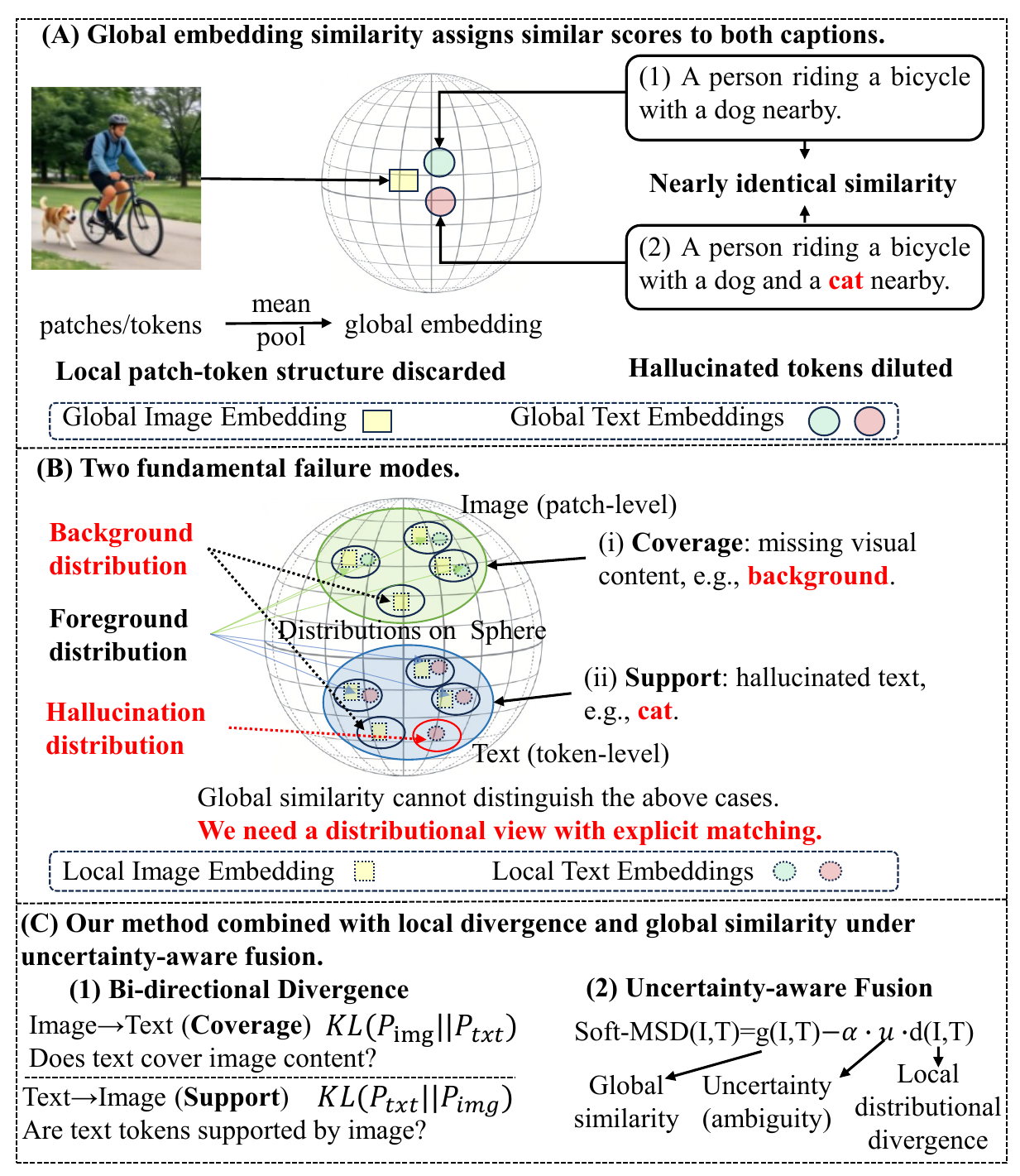}
    \vspace{-4mm}
    \caption{ \textbf{From pointwise similarity to distributional alignment.} (a) Global similarity methods (e.g. CLIPScore  \cite{hessel2021clipscore}) encode an image and a caption into single vectors via mean pooling and compute their similarity. Hallucinated content (e.g., ``a cat'') is diluted by dominant semantics, yielding similar scores to correct captions. (b) This is because mean pooling causes information collapse, discarding patch-token correspondences. It leads to two fundamental failure modes: (i) \textit{coverage}: some visual content is not described: (ii) \textit{support}: some text tokens lack visual grounding. (c) MSD-Score models image patches and text tokens as distributions on the unit hypersphere and measures alignment via bi-directional KL divergence to capture coverage and support. Combined with global similarity under uncertainty-aware fusion, MSD-Score provides a principled and fine-grained evaluation of image captions.
    }
    \vspace{-4mm}
    \label{fig:motivation}
\end{figure}

\IEEEPARstart{I}{mage} caption evaluation is a fundamental problem in vision--language research, serving as the primary tool for benchmarking captioning systems and guiding progress in multimodal understanding. A high-quality caption should faithfully reflect the visual content of an image, accurately describing salient objects, attributes, and relations while avoiding unsupported or hallucinated details \cite{kaul2024throne}. However, as modern vision--language models (VLMs) scale, they increasingly produce rich, open-ended descriptions that extend beyond the limited scope of human-annotated references. This creates a growing mismatch between open-ended visual semantics and reference-based evaluation protocols. Most existing benchmarks, such as COCO Captions \cite{chen2015microsoft}, Flickr30k \cite{young2014image}, and Visual Genome \cite{krishna2017visual}, rely on a small set of reference captions per image. While indispensable for supervised training, these references provide sparse and incomplete semantic coverage. As a result, captions that are factually correct but semantically different from the references are often unfairly penalized, motivating increasing interest in \emph{reference-free} caption evaluation methods that assess image--text alignment directly.

\begin{figure*}[!t]
    \centering
    \includegraphics[width=\textwidth]{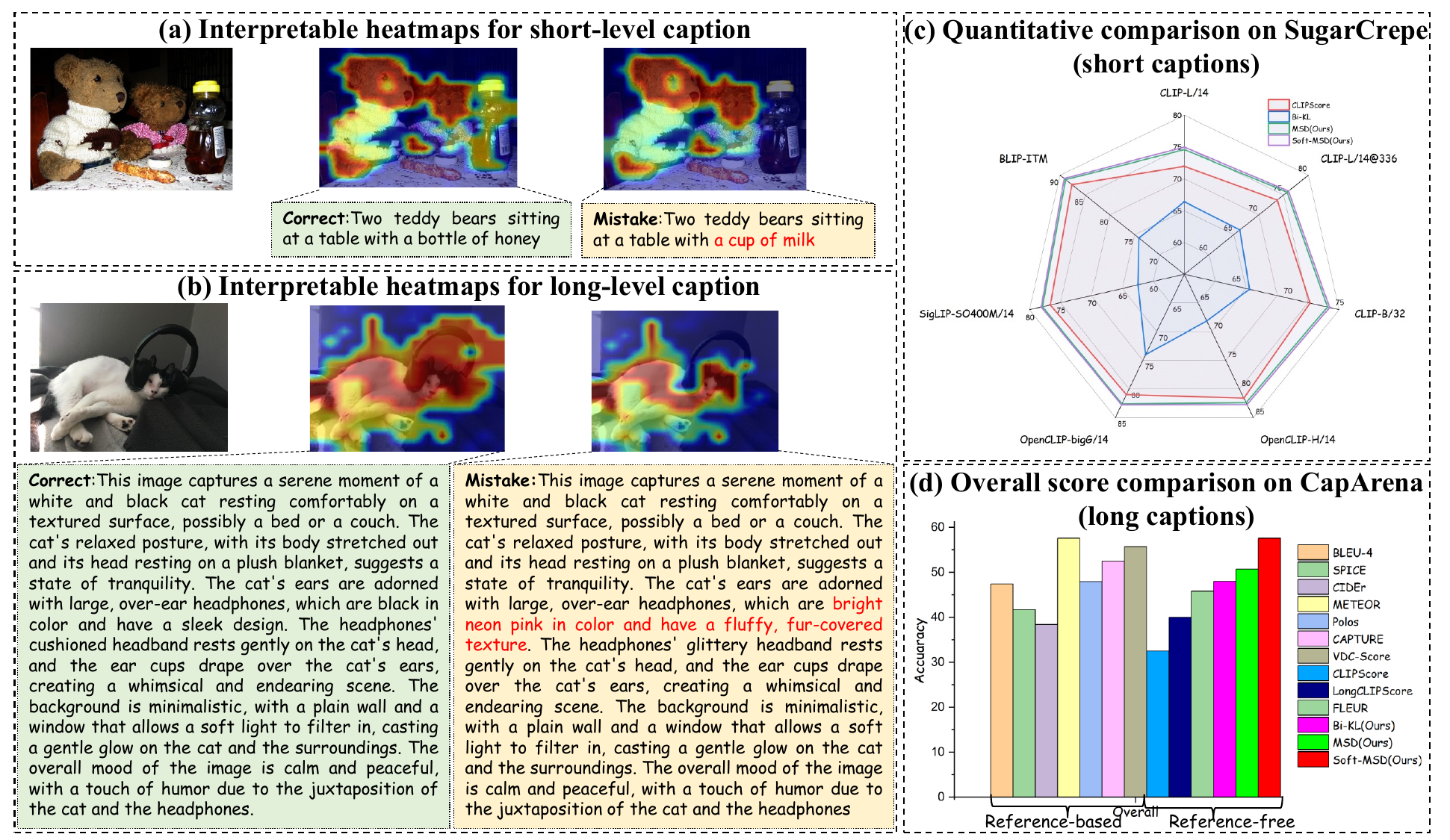}
    \vspace{-4mm}
    \caption{\textbf{Interpretability and benchmark evaluation of MSD-Score across short and long captions.} Panels (a) and (b) show KL-decomposition-based heatmaps for short- and long-form captions. MSD-Score localizes hallucinated tokens (red text, e.g., ``a cup of milk'', ``bright neon pink in color and have a fluffy, fur-covered texture'') and unsupported details; panels (c) and (d) summarize benchmark behavior on SugarCrepe \cite{hsieh2023sugarcrepe} and CapArena \cite{cheng2025caparena}, respectively. MSD-Score consistently outperforms strong reference-free baselines.
    }
    \label{fig:inter-results}
    \vspace{-4mm}
\end{figure*}

A common reference-free strategy embeds images and captions into a shared representation space and scores their alignment using global similarity. Representative methods include CLIPScore \cite{hessel2021clipscore}, LongCLIPScore \cite{zhang2024long}, and FLEUR \cite{lee2024fleur}. Throughout this paper, we use \emph{global} alignment to denote the holistic pooled alignment signal produced by mean-pooled image and text embeddings, and \emph{local} alignment to denote patch/token-level representations in a shared semantic space without explicit hard matching. As illustrated in Fig.~\ref{fig:motivation}(a), global-only approaches summarize alignment with a single similarity score computed from pooled embeddings. While effective for coarse relevance, this paradigm systematically discards local semantic structure. By collapsing patch-level visual features and token-level textual features into global vectors, fine-grained mismatches, such as missing attributes, incorrect relations, or hallucinated entities, are diluted by dominant correct content, allowing locally incorrect but globally plausible captions to receive high scores. Recent VLM-as-a-Judge methods \cite{hurst2024gpt} attempt to address this issue via holistic reasoning, but typically operate as uncalibrated black boxes, offering limited interpretability and sensitivity to prompt design. These limitations suggest that image--text alignment is inherently \emph{multi-scale} and \emph{distributional}: multi-scale here refers to semantic granularity rather than network depth. From this perspective, a caption is semantically faithful if its token-level concepts are well \emph{supported} by visual evidence, and if the visual content is sufficiently \emph{covered} by the textual description (Fig.~\ref{fig:motivation}(b)). The support term captures hallucinated or otherwise unsupported textual content, while the coverage term captures visual content that remains insufficiently described. Capturing these complementary properties requires explicit comparison between image and text distributions, rather than scalar similarity scores.

Motivated by this view, we propose \textbf{MSD-Score}, a reference-free image caption evaluation metric that formulates caption evaluation as a multi-scale distributional alignment problem. Given locally aligned patch-level and token-level embeddings, we model each modality as a probabilistic distribution over latent semantic directions using generative mixture models on the unit hypersphere. To quantify discrepancies between image and text distributions, we introduce a weighted bi-directional Kullback--Leibler divergence that explicitly captures complementary semantic failure modes corresponding to visual omissions and textual hallucinations. Finally, we integrate this local, distribution-level verification with global image--text similarity in a unified multi-scale framework, where an uncertainty-aware fusion mechanism adaptively modulates the contribution of local evidence (Fig.~\ref{fig:motivation}(c)). MSD-Score is not intended to be the lightest zero-shot metric or a replacement for judge-based evaluators (e.g., LLM-as-a-Judge). Instead, it targets higher-fidelity, yet tractable offline evaluation by combining deterministic local verification with standard global similarity.

Extensive experiments across diverse benchmarks demonstrate that MSD-Score achieves the best overall evaluation performance among reference-free metrics. As shown in Fig.~\ref{fig:inter-results}(c,d), MSD-Score performs competitively on both short-caption evaluation (SugarCrepe \cite{hsieh2023sugarcrepe}) and long-caption evaluation (CapArena \cite{cheng2025caparena}). Beyond numerical performance, MSD-Score provides interpretable diagnostics by decomposing distributional divergence into region- and token-level contributions, producing intuitive heatmaps that highlight hallucinated or unsupported content in both short- and long-form captions (Fig.~\ref{fig:inter-results}(a,b)). Overall, MSD-Score offers a principled, deterministic, and decomposable complementary signal to holistic similarity metrics and judge-based evaluators.

Our contributions are summarized as follows:
\begin{itemize}
\item We reformulate reference-free image caption evaluation as a \emph{multi-scale distributional alignment} problem, capturing fine-grained semantic discrepancies missed by global embedding-based metrics.
\item We propose a probabilistic framework that models image patches and text tokens as von Mises--Fisher mixtures on the unit hypersphere, and quantify alignment via a weighted bi-directional KL divergence sensitive to complementary failure modes.
\item We develop a multi-scale scoring mechanism that integrates global similarity with local distributional verification under uncertainty-aware fusion.
\item Extensive experiments show that MSD-Score achieves state-of-the-art alignment with human judgments across multiple benchmarks, while providing transparent diagnostics of local and global grounding errors.
\end{itemize}

\section{Related Work}

\noindent\textbf{Reference-based metrics.}
Traditional image caption evaluation has relied heavily on human-annotated references. Early approaches measure n-gram overlap, including BLEU~\cite{papineni2002bleu}, METEOR~\cite{banerjee2005meteor}, ROUGE~\cite{lin2004rouge}, and CIDEr~\cite{vedantam2015cider}, while SPICE~\cite{anderson2016spice} additionally exploits scene graph structures to capture semantic relations. More recently, embedding-augmented metrics such as RefCLIPScore~\cite{hessel2021clipscore}, CAPTURE~\cite{dong2024benchmarking}, and VDC-Score~\cite{chai2024auroracap} incorporate pre-trained vision-language representations to improve sensitivity to semantic similarity. Despite their improvements, these methods remain fundamentally constrained by the coverage and diversity of human references, often failing to reward captions with valid but unseen content.

\noindent\textbf{Reference-free metrics and token/region-level alignment.}
To overcome the limitations of reference dependency, reference-free metrics assess image--text alignment directly in pre-trained multimodal spaces. CLIPScore~\cite{hessel2021clipscore}, LongCLIPScore~\cite{zhang2024long}, and FLEUR~\cite{lee2024fleur} compute holistic similarity between image and text embeddings. Recent reference-free metrics also strengthen the evaluation space or introduce finer-grained cues, including UMIC~\cite{lee2021umic}, PAC-S~\cite{sarto2023pacs}, InfoMetIC~\cite{hu2023infometic}, BRIDGE~\cite{sarto2024bridge}, HICE-S~\cite{zeng2024hicescore}, and EXPERT~\cite{kim2025expert}. Broadly, these methods either learn caption-quality predictors from supervision, enrich the visual/textual cues used at evaluation time, or provide structured feedback about caption quality. MSD-Score is complementary to this line of work: instead of scoring alignment with a pointwise predictor, it models image patches and text tokens as local distributions and compares them through an explicit probabilistic discrepancy.

\noindent\textbf{Distributional and probabilistic metrics.}
An emerging line of work treats images and texts as distributions over latent embeddings rather than single points. Optimal Transport-based metrics, such as Word Mover's Distance~\cite{kusner2015word} and Sinkhorn distances~\cite{cuturi2013sinkhorn}, compare sets of embeddings through principled distributional similarity. Gaussian mixture embeddings have been explored in semantic matching and anomaly localization~\cite{defard2021padim, liang2022gmmseg}, capturing multi-component structures in high-dimensional feature spaces. These distributional approaches inspire MSD-Score: we explicitly model patch-level and token-level embeddings as von Mises--Fisher mixtures, leverage the spherical geometry of normalized vision-language embeddings, decompose discrepancies into coverage and support, and integrate the resulting local signal with global similarity via uncertainty-aware fusion.

\noindent\textbf{LLM-as-a-Judge.}
More recently, large language models have been employed as evaluators of visual content. GPT-4o~\cite{hurst2024gpt}, open-source models like Qwen2~\cite{wang2024qwen2}, and LLaVA-style judges~\cite{li2024llava, xiong2025llava} demonstrate impressive heuristic performance in judging caption quality. G-Eval~\cite{liu2023g} further formalizes LLM evaluation for multi-modal tasks. Nevertheless, these models act as uncalibrated black boxes: their judgments are sensitive to prompt design and lack explicit probabilistic interpretation, limiting their reliability for fine-grained error analysis. MSD-Score is therefore positioned as a deterministic, decomposable, and complementary signal rather than as a replacement for judge-based evaluation.

\section{Method}

\begin{figure*}[!t]
    \centering
    \includegraphics[width=\textwidth]{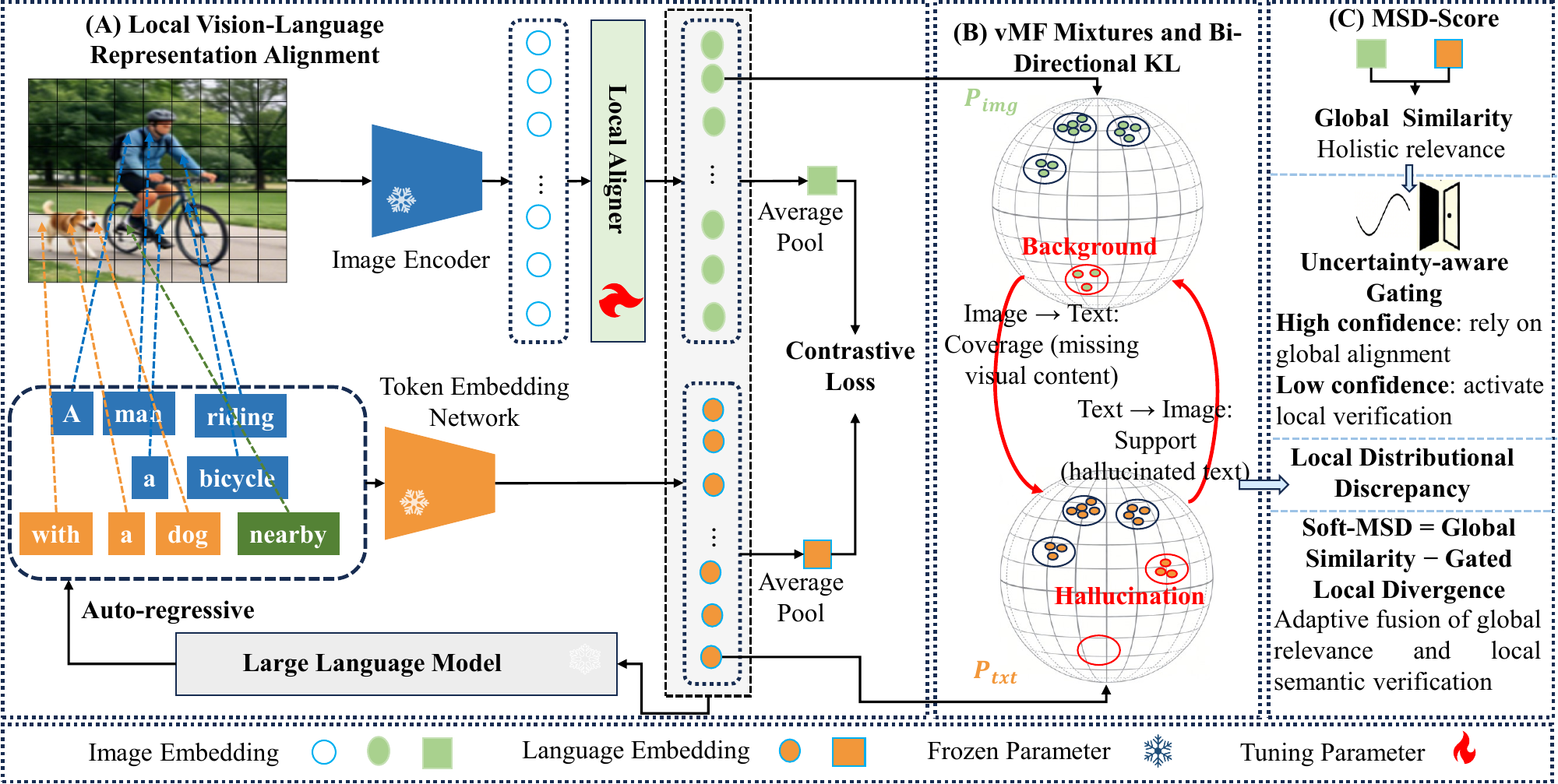}
    \vspace{-4mm}
    \caption{\textbf{Overview of the proposed MSD-Score framework for uncertainty-aware vision–language discrepancy modeling.} We introduce a multi-scale scoring paradigm that unifies global alignment with local distributional verification. (A) A frozen CLIP encoder and LLaMA model provide patch- and token-level embeddings, while a learnable alignment module maps visual patches into a shared semantic space and is trained with contrastive and reconstruction objectives. (B) Aligned representations are modeled as von Mises–Fisher mixtures, enabling bidirectional KL divergence to quantify missing visual evidence and textual hallucinations. (C) A confidence-aware gating mechanism adaptively combines global similarity with local discrepancy, producing the final Soft-MSD score. The LLM is used only during training as a frozen regularizer and removed at inference, ensuring efficient and backbone-agnostic scoring.
    }
    \label{fig:framwork_overview}
    \vspace{-4mm}
\end{figure*}

In this section, we introduce a unified framework for evaluating image caption quality by assessing semantic alignment between visual and textual representations through multi-scale distributional matching. As illustrated in Fig.~\ref{fig:framwork_overview}, our approach differs fundamentally from global similarity-based metrics such as CLIPScore~\cite{hessel2021clipscore}, which compress each modality into a single pooled embedding and measure alignment using cosine similarity. Such holistic pooling often obscures fine-grained semantic discrepancies. Instead, we preserve patch-level and token-level structures by modeling the full sets of aligned visual and textual embeddings as probability distributions and measuring their discrepancy at the distribution level. This formulation enables fine-grained sensitivity to subtle semantic errors, including missing attributes, incorrect relations, and hallucinated entities, which are difficult to capture with global similarity alone.

As shown in Fig.~\ref{fig:framwork_overview}, the proposed framework consists of four interconnected components: (1) a local vision--language alignment module that produces semantically aligned image patch and text token embeddings (Fig.~\ref{fig:framwork_overview} (A); Sec.~\ref{sec:token_alignment});
(2) distributional modeling of local embeddings using von Mises--Fisher mixture models on the unit hypersphere (Fig.~\ref{fig:framwork_overview} (B); Sec.~\ref{sec:vmf_modeling});
(3) bi-directional distributional divergence to capture complementary semantic discrepancies corresponding to omissions and hallucinations (Fig.~\ref{fig:framwork_overview} (B); Sec.~\ref{sec:divergence}); and
(4) a multi-scale aggregation mechanism that integrates local divergence with global similarity via uncertainty-aware fusion to produce the final MSD and Soft-MSD scores (Fig.~\ref{fig:framwork_overview} (C); Sec.~\ref{sec:fusion}).

\subsection{Problem Formulation}
\label{sec:problem}

Given an image and a candidate caption, our goal is to assess semantic alignment at both global and local levels. Here, global alignment refers to the holistic pooled image--text signal, while local alignment refers to patch/token-level representations in a shared semantic space. Modern vision-language models such as CLIP~\cite{radford2021learning-clip} and SigLIP~\cite{tschannen2025siglip} excel at image-sentence alignment but primarily focus on global representations, providing limited supervision for fine-grained, token-level correspondences. To address this, we train a local vision-language alignment model (Sec.~\ref{sec:token_alignment}) that produces patch-level visual embeddings and token-level textual embeddings. These embeddings capture fine-grained attributes, object parts, and relational semantics, forming the foundation for distribution-level modeling. Formally, let
\begin{equation}
\mathcal{X}_{\text{img}} = \{ \mathbf{x}_1, \dots, \mathbf{x}_{N_{\text{img}}} \}, \quad
\mathcal{Y}_{\text{txt}} = \{ \mathbf{y}_1, \dots, \mathbf{y}_{N_{\text{txt}}} \}, \nonumber
\end{equation}
where $\mathbf{x}_i, \mathbf{y}_j \in \mathbb{R}^D$ are the (normalized) representations of visual patches and text tokens. Typically, $N_{\text{img}} \approx 20$--$50$ and $N_{\text{txt}} \approx 5$--$40$, with strong internal structure. This is a high-dimensional, low-sample regime: each instance contains only tens of local elements, while the embedding dimension is substantially larger.

We hypothesize that an image's semantics are better represented as a \emph{distribution} over latent directions rather than a single point in $\mathbb{R}^D$, reflecting visual attributes, relations, and fine-grained details. A high-quality caption should induce a distribution whose modes and variances closely match those of the image. To operationalize this, we fit a $K$-component generative mixture model to each embedding set:
\begin{align}
    P_{\text{img}}(\mathbf{x}) &= \sum_{k=1}^{K_{\text{img}}} \pi^{\text{img}}_k \, p_k(\mathbf{x}; \boldsymbol{\theta}^{\text{img}}_k), \\
    P_{\text{txt}}(\mathbf{y}) &= \sum_{k=1}^{K_{\text{txt}}} \pi^{\text{txt}}_k \, p_k(\mathbf{y}; \boldsymbol{\theta}^{\text{txt}}_k).
\end{align}
where $\pi_k^{(\cdot)}$ are mixture weights ($\sum_k \pi_k = 1$) and $p_k(\cdot; \boldsymbol{\theta}_k)$ is the $k$-th component density (Sec.~\ref{sec:vmf_modeling}). Under this formulation, semantic alignment reduces to measuring the discrepancy between two high-dimensional distributions:
\begin{equation}
\mathcal{D}\bigl(P_{\text{img}} \,\|\, P_{\text{txt}}\bigr),
\end{equation}
where $\mathcal{D}(\cdot \| \cdot)$ is a suitable distributional divergence measure (Sec.~\ref{sec:divergence}).

\subsection{Local Vision-Language Representation Alignment}
\label{sec:token_alignment}

\begin{figure*}[!t]
  \centering
  \begin{tabular}{cc}
    \textbf{Contrastive only} & \textbf{Contrastive + reconstruction} \\
    \includegraphics[width=0.48\textwidth]{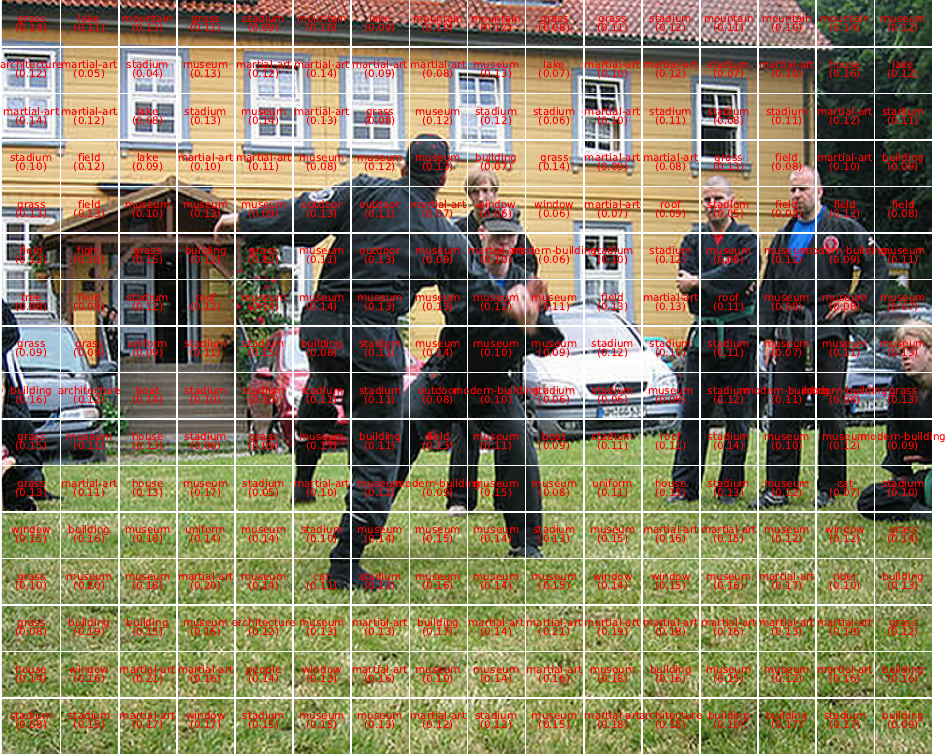} & \includegraphics[width=0.48\textwidth]{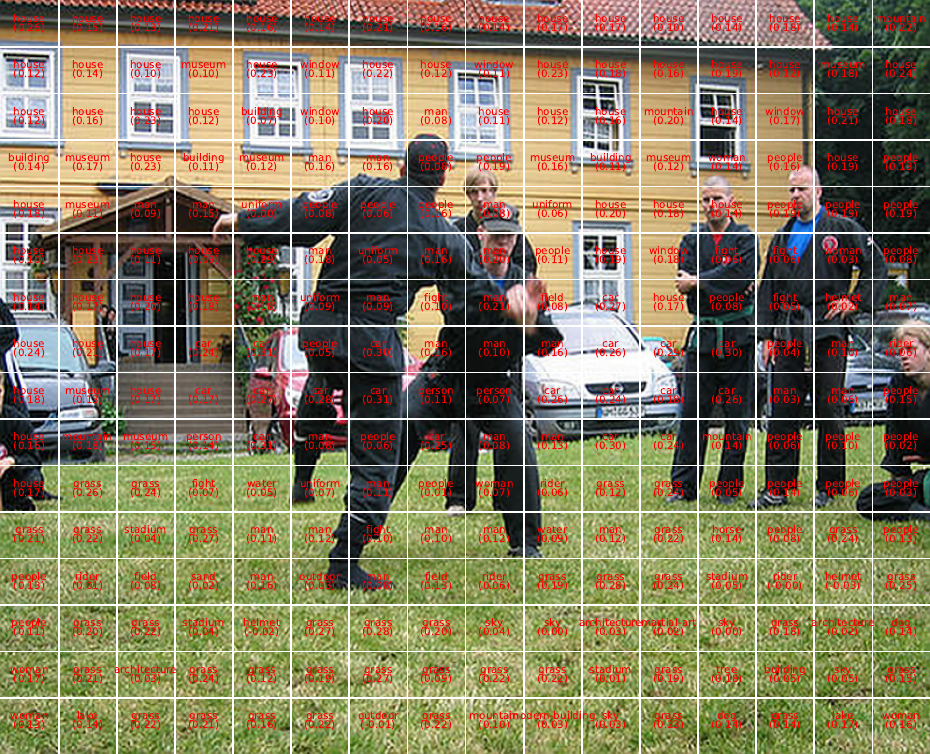} \\
  \end{tabular}
  \vspace{-4mm}
 \caption{Effect of the reconstruction objective on local semantic alignment. 
We compare a contrastive-only aligner with our full contrastive–generative alignment framework. Adding reconstruction produces more concentrated and semantically consistent local responses, leading to improved token-level structure for downstream discrepancy modeling. }
  \label{fig:recon_qual_1}
  \vspace{-4mm}
\end{figure*}

\begin{table*}[!ht]
  \centering
  \renewcommand{\arraystretch}{1.15}
  \caption{Analysis of the reconstruction objective under identical backbone, data, and scoring pipeline. Removing reconstruction consistently degrades performance across benchmarks, indicating that it improves fine-grained semantic sensitivity.}
\label{tab:reconstruction_ablation}
  \label{tab:recon_ablation}
  \resizebox{0.7\textwidth}{!}{
  \begin{tabular}{l|ccc|ccc}
    \thickhline
    \multirow{2}{*}{\textbf{Variant}} & \multicolumn{3}{c|}{\textbf{DocENT} \cite{ananthram2025posh}} & \multicolumn{3}{c}{\textbf{CapArena} \cite{cheng2025caparena}} \\
     & Overall & Spearman & Kendall & Overall & Spearman & Kendall \\
    \hline
    CLIPScore & 53.5 & 0.181 & 0.136 & 32.5 & $-0.574$ & $-0.451$ \\
    w/o reconstruction & 60.8 & 0.224 & 0.163 & 54.0 & 0.342 & 0.263 \\
    \textbf{w/ reconstruction} & \textbf{64.8} & \textbf{0.259} & \textbf{0.195} & \textbf{57.6} & \textbf{0.398} & \textbf{0.297} \\
    \thickhline
  \end{tabular}}
  \vspace{-4mm}
\end{table*}

To model image captions via local embedding distributions, we require a vision-language encoder that produces semantically aligned image patches and text tokens, rather than relying on global pooling alone. Instead of retraining a full VLM, we learn a lightweight local aligner on frozen backbones to obtain fine-grained patch- and token-level representations. We introduce a \emph{contrastive generative alignment} framework (Fig.~\ref{fig:framwork_overview}(A)) that jointly optimizes: (1) a contrastive objective enforcing global image–text consistency after pooling, and (2) an auto-regressive reconstruction objective that regularizes token-level semantics for caption generation. Formally, let
\begin{equation}
\bar{\mathbf{v}}=\mathbb{E}[\mathcal{X}_{\text{img}}],\quad 
\bar{\mathbf{u}}=\mathbb{E}[\mathcal{Y}_{\text{txt}}], \nonumber
\end{equation}
denote the $\ell_2$-normalized global representations. The symmetric contrastive loss is
\begin{equation}
\begin{aligned}
\mathcal{L}_{\text{contra}} 
= & -\frac{1}{2B} \sum_{i=1}^{B} 
\Bigg[
\log \frac{\exp(\bar{\mathbf{v}}_i^\top \bar{\mathbf{u}}_i / \tau)}
{\sum_{j=1}^{B} \exp(\bar{\mathbf{v}}_i^\top \bar{\mathbf{u}}_j / \tau)} \\
& + 
\log \frac{\exp(\bar{\mathbf{u}}_i^\top \bar{\mathbf{v}}_i / \tau)}
{\sum_{j=1}^{B} \exp(\bar{\mathbf{u}}_i^\top \bar{\mathbf{v}}_j / \tau)}
\Bigg],
\end{aligned}
\end{equation}
where $\tau$ is a temperature and $B$ is the batch size.

To further regularize fine-grained semantics, image patches are projected into the embedding space of a frozen LLM as a continuous visual prefix, followed by caption tokens, and the model is trained to reconstruct the caption in an auto-regressive manner:
\begin{equation}
q(\bm{Y}_T \mid \bm{X}_I) = \prod_{i=1}^{L} q_{\bm{\theta}}(\bm{y}_i \mid \bm{X}_I, \bm{Y}_{T,<i}),
\end{equation}
where $\bm{\theta}$ denotes the trainable parameters and $\bm{Y}_{T,<i}$ represents preceding tokens. The objective is the negative log-likelihood of the ground-truth caption. Importantly, the LLM is used exclusively as a training-time semantic regularizer and is discarded during inference, ensuring that no additional computational overhead is introduced at test time. At scoring stage, our algorithm operates purely on the learned patch- and token-level embeddings to produce a more reliable discrepancy estimation under the same backbone setting.

Intuitively, contrastive learning enforces global image–text alignment but remains insufficient to capture token-level structure, whereas reconstruction introduces an explicit semantic constraint that improves local consistency modeling. Joint optimization of both objectives produces structured token-level embeddings that serve as a foundation for robust local distribution modeling. Here, \emph{alignment} refers to locally paired vision–language representations in a shared semantic space, rather than explicit hard matching. As illustrated in Figure~\ref{fig:recon_qual_1}, introducing the reconstruction objective leads to more concentrated and semantically consistent local activation patterns, indicating improved token-level structure. This effect is quantitatively supported in Table~\ref{tab:recon_ablation}, where removing reconstruction results in consistent performance degradation across benchmarks (e.g., from 64.8 to 60.8 in total in DocENT \cite{ananthram2025posh}, and from 57.6 to 54.0 in CapArena \cite{cheng2025caparena}), demonstrating its critical role in enhancing fine-grained semantic sensitivity rather than serving as an auxiliary inference module.

\subsection{Distributional Modeling of Local Vision-Language Embeddings}
\label{sec:vmf_modeling}

Given locally aligned patch- and token-level embeddings (Sec.~\ref{sec:token_alignment}), we aim to model fine-grained vision--language correspondences as structured distributions over semantic directions. As illustrated in Figure~\ref{fig:framwork_overview}(B), we formulate each modality as a mixture distribution on the unit hypersphere, enabling principled modeling of multi-modal local semantics. Since both patch and token embeddings are $\ell_2$-normalized, they lie naturally on the hypersphere $\mathbb{S}^{D-1}$. This induces a geometric structure where similarity is better characterized by angles rather than Euclidean distances. Motivated by this property, we model local embeddings using a \emph{von Mises--Fisher (vMF) mixture}, which operates directly on directional data and aligns with cosine-based similarity used in contrastive vision--language learning. In contrast, Gaussian mixture models (GMMs) assume Euclidean geometry and rely on covariance estimation, which becomes ill-posed in high-dimensional and few-token regimes. This mismatch leads to suboptimal modeling of semantic directions.

\begin{table}[!t]
  \centering
  \renewcommand{\arraystretch}{1.25}
   \caption{Analysis on local distribution modeling choices (pairwise accuracy \%) on the COCO-CF dataset. We compare Gaussian mixture models (GMM, Euclidean) with adaptive-$\kappa$ vMF and fixed-$\kappa=20$ vMF (ours). Fixed-$\kappa$ vMF consistently achieves the highest performance across all compositional error types, emphasizing the benefit of stable concentration control and spherical modeling.}
  \label{tab:ablation_kappa_gmm}
  \resizebox{\linewidth}{!}{%
  \begin{tabular}{l|ccc}
    \thickhline
    \textbf{Subtask} 
    & \textbf{GMM (Euclidean)} 
    & \textbf{vMF (Adaptive $\kappa$)} 
    & \textbf{vMF (Fixed $\kappa=20$, Ours)} \\
    \hline
    Add-Att     & 60.40 & 75.45 & \textbf{76.88} \\
    Add-Obj     & 67.85 & 77.35 & \textbf{79.39} \\
    Replace-Att & 77.41 & 76.14 & \textbf{80.33} \\
    Replace-Obj & 93.34 & 92.49 & \textbf{94.07} \\
    Replace-Rel & 58.04 & 53.65 & \textbf{66.36} \\
    Swap-Att    & 60.66 & 61.26 & \textbf{64.26} \\
    Swap-Obj    & 54.29 & 56.33 & \textbf{63.67} \\
    \thickhline
  \end{tabular}%
  }
  \vspace{-4mm}
\end{table}

\begin{figure}[!t]
  \centering
  \includegraphics[width=\linewidth]{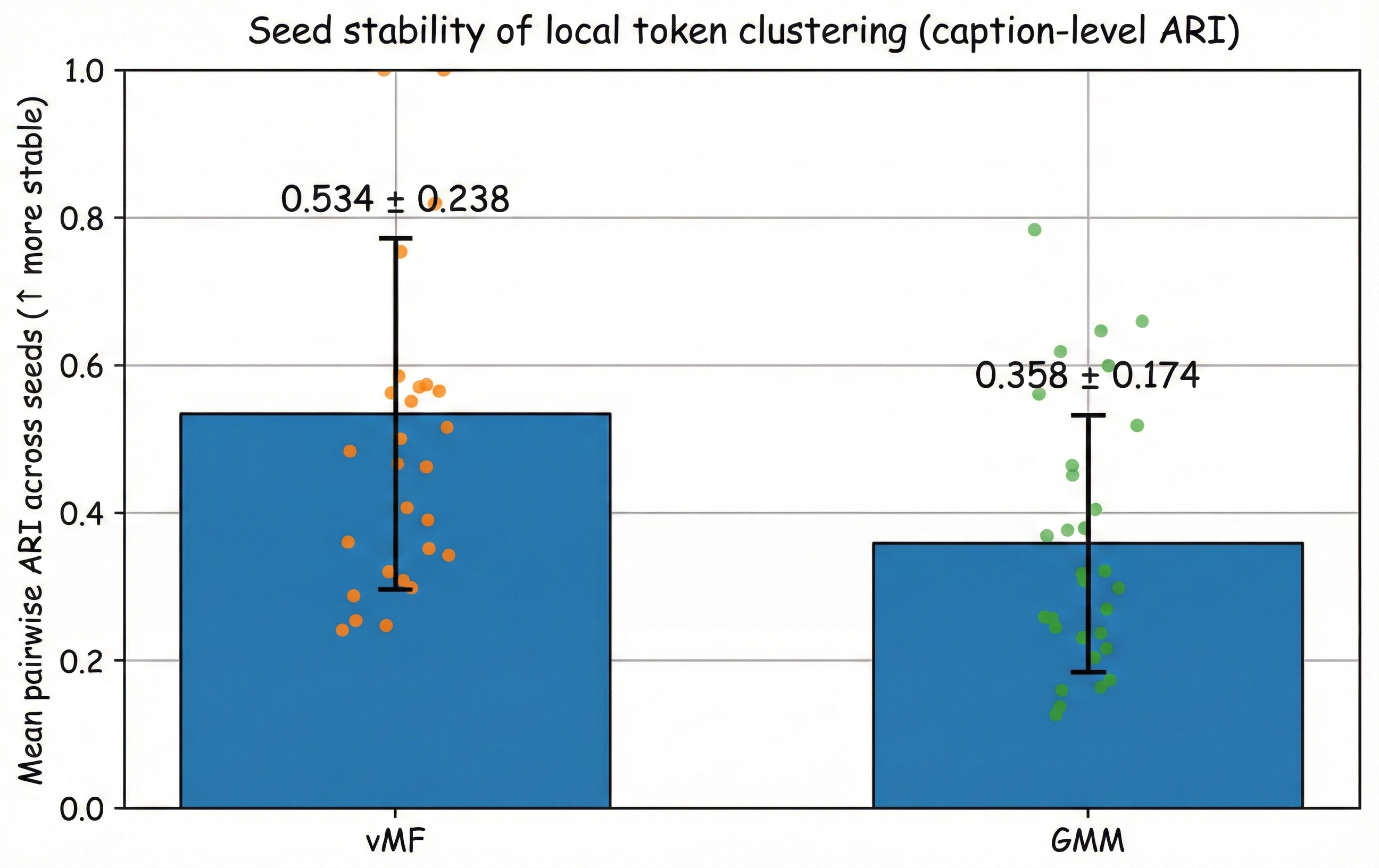}
  \vspace{-4mm}
  \caption{Seed stability of local token clustering (caption-level ARI).
  Each dot corresponds to one caption.
  For each caption, we run mixture fitting with multiple random seeds and compute the mean pairwise ARI between the resulting token clusterings.
  vMF yields more stable clusterings than Euclidean GMM in the high-dimensional, few-token regime.}
  \label{fig:seed_stability_ari}
  \vspace{-4mm}
\end{figure}

We empirically validate this design choice in Table~\ref{tab:ablation_kappa_gmm}. Across all compositional perturbations in COCO-CF, GMMs consistently underperform vMF-based models, particularly under additive and swap-based distortions (e.g., Swap-Obj: 54.29 vs. 63.67). This performance gap indicates that Euclidean assumptions fail to capture the angular semantic structure, especially when local token sets are few and high-dimensional. In contrast, vMF mixtures directly model angular similarity and avoid covariance estimation, making them significantly more robust under compositional variations. Beyond precision, we further analyze the stability of local distribution estimation under random initialization. As shown in Figure~\ref{fig:seed_stability_ari}, we compute the Adjusted Rand Index (ARI) for multiple EM (Sec.~\ref{sec:emsec}) runs with different seeds. vMF mixtures achieve substantially higher stability than GMMs (0.534$\pm$0.238 vs.\ 0.358$\pm$0.174), indicating more consistent semantic partitioning. This improvement demonstrates that spherical modeling not only improves accuracy but also yields a more reproducible latent structure in the few-token regime. Overall, these results justify our choice of vMF mixtures as a geometrically consistent and empirically stable distributional model for local vision--language embeddings.

For a unit vector $\mathbf{x} \in \mathbb{S}^{D-1}$, the vMF density is defined as
\begin{equation}
\begin{aligned} \mathrm{vMF}(\mathbf{x} \mid \boldsymbol{\mu}_k, \kappa_k) &= C_D(\kappa_k)\exp\bigl(\kappa_k \boldsymbol{\mu}_k^\top \mathbf{x}\bigr), \\ \|\boldsymbol{\mu}_k\|_2 &= 1, \quad \kappa_k > 0 , \end{aligned} 
\label{eq:vmf} 
\end{equation}
where $\boldsymbol{\mu}_k$ is the mean direction, $\kappa_k$ controls concentration, and $C_D(\kappa_k)$ is a normalization constant. Inner products $\boldsymbol{\mu}_k^\top \mathbf{x}$ correspond to cosine similarity, aligning naturally with contrastive vision-language embeddings. To capture heterogeneous semantics, we model each modality as a $K$-component mixture:
\begin{equation}
P(\mathbf{x}) = \sum_{k=1}^{K} \pi_k\, C_D(\kappa_k) \exp(\kappa_k \boldsymbol{\mu}_k^\top \mathbf{x}), 
\quad \sum_{k=1}^{K} \pi_k = 1,
\label{eq:vmf_mix}
\end{equation}
where each component $(\boldsymbol{\mu}_k, \kappa_k, \pi_k)$ acts as a latent semantic prototype (objects, attributes, relations). One mixture is fitted to image patch embeddings $P_{\text{img}}$, and another to text token embeddings $P_{\text{txt}}$. Parameters are estimated via the EM algorithm (Algorithm~\ref{alg:vmf_em}), which alternates between the soft assignment of tokens to mixture components (E-step) and updating weights of the mixture and the mean directions (M-step). In high-dimensional, few-token regimes, reliable estimation of mixture parameters becomes challenging due to limited observations per component. 

To improve numerical stability, we fix the concentration parameter $\kappa$ in all components and omit the normalization constant during optimization. This imposes a shared angular bandwidth and reduces variance in local distribution fitting. We empirically validate this design choice in Table~\ref{tab:ablation_kappa_gmm}. Fixed-$\kappa$ vMF consistently outperforms both adaptive-$\kappa$ vMF and Euclidean GMMs across all compositional perturbations on COCO-CF, achieving gains of +8.32 on Replace-Rel and +7.34 on Swap-Obj over GMM.  We attribute this improvement to the underdetermined nature of token-level clustering: allowing $\kappa$ to vary introduces additional degrees of freedom, which amplifies estimation variance and leads to unstable semantic partitioning. In contrast, fixing $\kappa$ enforces a uniform angular resolution across components, resulting in more stable and generalizable local semantic structures. These results demonstrate that fixed-$\kappa$ vMF provides a geometrically consistent and statistically stable foundation for modeling local vision--language distributions, which is crucial for reliable bidirectional discrepancy estimation in our framework.

\subsection{Parameter Estimation for vMF Mixtures}
\label{sec:emsec}

In this section, we describe how the parameters of the vMF mixture model are estimated from a finite set of locally aligned token embeddings.
Given a set of observations, our objective is to maximize the log-likelihood of the data under the proposed mixture model with respect to the mixture weights $\{\pi_k\}_{k=1}^K$ and the mean directions $\{\boldsymbol{\mu}_k\}_{k=1}^K$. Direct maximization of this objective is challenging due to the presence of latent component assignments, which couple the mixture weights and component parameters. To address this, we introduce latent assignment variables and adopt an Expectation-Maximization (EM) framework, which provides a principled and well-established approach for maximum likelihood estimation in mixture models. The EM procedure alternates between estimating soft assignment probabilities for each token embedding (E-step) and updating the corresponding semantic prototypes and mixture weights (M-step), monotonically increasing the data likelihood at each iteration.

In practice, token-level representations exhibit a highly underdetermined regime, characterized by high embedding dimensionality (e.g., $D \approx 768$) and a limited number of samples per instance (typically tens of tokens).
Under such conditions, jointly estimating all the parameters of the mixture can lead to unstable solutions and a high variance of the estimation. In the above subsection, we fix the concentration parameter $\kappa_k = \kappa$ to improve numerical stability and ensure robust parameter estimation. When $\kappa$ is fixed, the vMF normalizing constant $C_D(\kappa_k)$ becomes independent of the weights of the mixture and the mean directions. Since it therefore does not affect the optimization of these parameters, we omit it during parameter estimation by setting $C_D(\kappa_k)=1$. A detailed discussion of the choice of $\kappa$ and the theoretical justification for canceling the normalization term is provided in subsection.~\ref{sec:cancel}.

Formally, we associate each token embedding $\mathbf{x}_i$ with a latent assignment variable $z_i \in \{1,\dots,K\}$, where $z_i = k$ indicates that $\mathbf{x}_i$ is assigned to the $k$-th mixture component. Based on these latent assignments, parameter estimation proceeds via alternating E- and M-steps. For numerical robustness, all computations are carried out in log-space using the log-sum-exp trick. We next detail the E-step and M-step used to optimize the vMF mixture model.

\noindent\textbf{E-step (responsibilities).}
In the E-step, we compute the posterior responsibilities of mixture components for each token embedding. With a shared concentration parameter $\kappa$, the responsibility of component $k$ for token $\mathbf{x}_i$ is given by
\begin{equation}
\gamma_{ik} := p(z_i = k \mid \mathbf{x}_i)
= \frac{\pi_k \exp\bigl(\kappa \boldsymbol{\mu}_k^\top \mathbf{x}_i\bigr)}
       {\sum_{j=1}^{K} \pi_j \exp\bigl(\kappa \boldsymbol{\mu}_j^\top \mathbf{x}_i\bigr)}.
\label{eq:estep}
\end{equation}
Here, $\gamma_{ik} \in [0,1]$ represents a soft assignment indicating the extent to which token $\mathbf{x}_i$ is explained by component $k$. Since both $\mathbf{x}_i$ and $\boldsymbol{\mu}_k$ are $\ell_2$-normalized, the inner product $\boldsymbol{\mu}_k^\top \mathbf{x}_i$ corresponds to cosine similarity, ensuring that responsibility estimation is consistent with the geometry of vision-language embeddings.

\noindent\textbf{M-step (mixture weights and mean directions).}
In the M-step, we update the mixture parameters by maximizing the expected complete-data log-likelihood under the current responsibilities.
Let
\begin{equation}
N_k = \sum_{i=1}^{N} \gamma_{ik}
\end{equation}
denote the effective number of tokens assigned to component $k$.
The parameter updates are given by
\begin{equation}
\pi_k \leftarrow \frac{N_k}{N}, \qquad
\boldsymbol{\mu}_k \leftarrow 
\frac{\sum_{i=1}^{N} \gamma_{ik} \mathbf{x}_i}
     {\left\|\sum_{i=1}^{N} \gamma_{ik} \mathbf{x}_i\right\|_2}.
\label{eq:mstep}
\end{equation}
The update for $\boldsymbol{\mu}_k$ corresponds to a responsibility-weighted resultant direction, followed by renormalization to enforce the unit-norm constraint $\boldsymbol{\mu}_k \in \mathbb{S}^{D-1}$. Each EM iteration therefore refines the component prototypes toward directions that best summarize the semantic structure of their assigned token embeddings.
The complete EM procedure is summarized in \textbf{Algorithm~\ref{alg:vmf_em}}.

\begin{algorithm}[t]
\caption{EM algorithm for parameter estimation of vMF mixtures}
\label{alg:vmf_em}
\begin{algorithmic}[1]
\REQUIRE Unit vectors $X=\{\mathbf{x}_i\}_{i=1}^{N}\subset\mathbb{S}^{D-1}$; number of components $K$; fixed $\kappa$; number of iterations $m$.
\ENSURE Mixture weights $\{\pi_k\}_{k=1}^{K}$ and mean directions $\{\boldsymbol{\mu}_k\}_{k=1}^{K}$.
\STATE Initialize $\boldsymbol{\mu}_k$ by sampling $K$ tokens from $X$ (with replacement if $N<K$), followed by $\ell_2$ normalization.
\STATE Initialize $\pi_k \leftarrow 1/K$.
\FOR{$t=1$ to $m$}
    \STATE \textbf{E-step:} compute responsibilities $\gamma_{ik}$ using Eq.~\eqref{eq:estep}.
    \STATE \textbf{M-step:} update $(\pi_k,\boldsymbol{\mu}_k)$ using Eq.~\eqref{eq:mstep}.
    \STATE Optionally reinitialize components with very small $N_k$.
\ENDFOR
\STATE \textbf{return} $\{\pi_k,\boldsymbol{\mu}_k\}_{k=1}^{K}$.
\end{algorithmic}
\end{algorithm}

This algorithm is performed at test time on a per-sample basis, without access to any training data. In this setting, particularly for limited text tokens, certain mixture components may receive extremely low or near-zero total responsibility. To address this issue, we introduce a simple reinitialization strategy. Specifically, if the effective component mass $N_k$ falls below a predefined threshold, we reinitialize its mean direction $\boldsymbol{\mu}_k$ using a randomly selected token from the current sample and reset the corresponding mixture weight. This prevents numerical degeneracy (e.g., vanishing normalization or unstable updates) while preserving the original probabilistic modeling objective.

\subsection{Stability and Invariance in Fixed-\texorpdfstring{$\kappa$}{kappa} vMF Modeling}
\label{sec:cancel}

The EM formulation in the above subsection adopts two deliberate simplifications: we treat the concentration parameter $\kappa$ as a fixed constant and omit the vMF normalization term $C_D(\kappa_k)$. While these choices significantly simplify estimation, they may appear restrictive at first glance. In this subsection, we provide a principled analysis showing that
both design decisions: (i) substantially improve numerical stability in high-dimensional, few-token regimes, and (ii) preserve all quantities relevant to EM inference and KL-based divergence computation.

Estimating the concentration parameter $\kappa$ from a limited number of tokens is notoriously unstable in high-dimensional spaces. Rather than adapting $\kappa$ per instance or per component, we adopt a \emph{fixed-$\kappa$} strategy and interpret $\kappa$ as a global angular bandwidth that controls the resolution at which directions in the embedding space are distinguished. This design avoids instance-wise overfitting and enforces a consistent notion of angular similarity across all mixtures, modalities, and samples.

In our setting, token distributions are fitted per instance (e.g., per image or per text), where the embedding dimension $D$ is high while the number of tokens $N_t$ is relatively small. Under standard vMF maximum likelihood estimation, $\kappa$ is updated based on the mean resultant length. For a single mixture component with soft assignments $\gamma_{ik}$, we define
\begin{equation}
\bar{\mathbf{s}}_k := \frac{1}{N_k}\sum_{i=1}^{N}\gamma_{ik}\mathbf{x}_i,
\qquad
\bar r_k := \|\bar{\mathbf{s}}_k\|_2 \in (0,1),
\label{eq:rbar}
\end{equation}
where $\bar{\mathbf{s}}_k$ is the responsibility-weighted mean direction
(before re-normalization), and $\bar r_k$ is the mean resultant length,
indicating how tightly assigned tokens concentrate around $\boldsymbol{\mu}_k$.
A common approximation (e.g., reference \cite{banerjee2005clustering}) maps $\bar r_k$ to an estimated concentration via
\begin{equation}
\hat{\kappa}(\bar r)
\approx
\frac{\bar r\,D-\bar r^3}{1-\bar r^2}.
\label{eq:kappa_approx}
\end{equation}
This mapping is monotone in $\bar r$: as $\bar r \to 1$, the estimated concentration grows rapidly, corresponding to an extremely peaked distribution.
However, this relationship is highly ill-conditioned. Taking the derivative yields
\begin{equation}
\frac{d\hat{\kappa}}{d\bar r}
=
\frac{D(1+\bar r^2)+\bar r^4-3\bar r^2}{(1-\bar r^2)^2}
=
\Theta\!\left(\frac{D}{(1-\bar r^2)^2}\right).
\label{eq:kappa_sensitivity}
\end{equation}
Eq.~\eqref{eq:kappa_sensitivity} reveals two compounding effects:
(i) the denominator $(1-\bar r^2)^2$ explodes as $\bar r \to 1$, and
(ii) the sensitivity scales linearly with the embedding dimension $D$. Consequently, in high-dimensional regimes, even small fluctuations in $\bar r_k$, which are inevitable when the effective cluster size $N_k$ is small,
can induce drastic swings in $\hat{\kappa}$. Within the EM procedure, an excessively large $\kappa$ produces overly confident posterior responsibilities in the E-step, which can lead to premature component collapse and unstable mixture fitting. This phenomenon is empirically illustrated in Figure~\ref{fig:resp_entropy_vs_length}. To mitigate these issues, we treat $\kappa$ as a global hyperparameter and fix it for all mixture components and all instances. This choice reduces the degrees of freedom of the model, acts as an implicit regularizer, and yields stable and reproducible divergence estimates. In practice, $\kappa$ is tuned once on a development split and kept fixed for all test samples.

The vMF normalization constant $C_D(\kappa_k)$ involves high-order modified Bessel functions and is numerically delicate to evaluate in high dimensions.
Fortunately, under the fixed-$\kappa$ design, this term can be safely omitted without affecting inference or divergence computation.

\noindent\textbf{(1) Invariance of EM Responsibilities.} Because $\kappa$ is shared across all components, $C_D(\kappa_k)$ is identical for every mixture term
and therefore cancels out in the E-step. As a result, responsibility updates in Eq.~\eqref{eq:estep} depend only on the inner products $\kappa\,\boldsymbol{\mu}_k^\top \mathbf{x}_i$ and the mixture weights $\pi_k$,
leaving the EM optimization unchanged.

\noindent\textbf{(2) Invariance of Directional KL Divergences.}
Write the log-density as
$\log p(\mathbf{x}) = \log C_D(\kappa_k) + \log \tilde p(\mathbf{x})$.
If both distributions $p_I$ and $p_t$
are constructed using the same fixed $\kappa$,
then
\begin{equation}
\log p_I(\mathbf{x}) - \log p_t(\mathbf{x})
=
\log \tilde p_I(\mathbf{x}) - \log \tilde p_t(\mathbf{x}),
\end{equation}
and the normalization constant cancels exactly. Therefore, all KL-based divergences remain unchanged, while numerical stability is substantially improved.

\subsection{Bi-directional Distributional Divergence for Semantic Discrepancy}
\label{sec:divergence}

Given the vMF mixtures $P_{\text{img}}$ and $P_{\text{txt}}$, we measure semantic discrepancy via a bi-directional KL divergence that captures two complementary error modes: \emph{omissions and hallucinations}. A fundamental property of image--text matching is that semantic errors are inherently asymmetric. When the caption is short, its limited capacity constrains the amount of information that can be expressed, making omissions (i.e., missing visual content) the dominant source of error. Conversely, when the caption is sufficiently long, it may exceed the visual evidence and introduce unsupported or hallucinated semantics, shifting the dominant error mode toward over-description. To model this asymmetry, we introduce a length-dependent weighting between the two KL directions:
\begin{equation}
d(I,T) = \beta(L) \, \mathrm{KL}(P_{\text{img}}\|P_{\text{txt}})
        + (1-\beta(L)) \, \mathrm{KL}(P_{\text{txt}}\|P_{\text{img}}),
\label{eq:weighted_div_refined}
\end{equation}
where $\beta(L) \in (0,1)$ balances omission and hallucination penalties. We define
\begin{equation}
\beta(L) = \frac{1}{1 + \exp\!\left(\frac{L - L_0}{\tau_L}\right)},
\end{equation}
where $L$ denote the caption length, $L_0$ denotes a reference caption length and $\tau_L$ controls the transition smoothness. As $L \to 0$, omission risk dominates the expected discrepancy, while as $L \to \infty$, hallucination risk becomes dominant. The proposed weighting $\beta(L)$ provides a smooth approximation to this asymmetric regime transition. This formulation yields a continuous shift from omission-sensitive evaluation (short captions) to hallucination-sensitive evaluation (long captions), avoiding brittle thresholding. It can be interpreted as a length-conditioned semantic risk estimator, where discrepancy is not treated as symmetric but adapts to the expressive capacity of the caption, which provides a principled bridge between global adequacy and local faithfulness, and serves as the foundation for robust multi-scale discrepancy scoring.

The two KL terms are defined as
\begin{align}
\mathrm{KL}(P_{\text{img}}\|P_{\text{txt}})
&= \mathbb{E}_{\mathbf{x}\sim P_{\text{img}}}[\log P_{\text{img}}(\mathbf{x}) - \log P_{\text{txt}}(\mathbf{x})], \nonumber \\
\mathrm{KL}(P_{\text{txt}}\|P_{\text{img}})
&= \mathbb{E}_{\mathbf{y}\sim P_{\text{txt}}}[\log P_{\text{txt}}(\mathbf{y}) - \log P_{\text{img}}(\mathbf{y})]. \nonumber
\end{align}
These correspond to two complementary semantic risks:
\begin{itemize}
\item \textbf{Coverage} ($\mathrm{KL}(P_{\text{img}} \,\|\, P_{\text{txt}})$): penalizes visual concepts present in the image but insufficiently captured by the text, reflecting omission errors.
\item \textbf{Support} ($\mathrm{KL}(P_{\text{txt}} \,\|\, P_{\text{img}})$): penalizes textual tokens that lack support in the image, reflecting hallucination errors.
\end{itemize}
Exact KL computation between mixtures is intractable, so we approximate expectations via Monte Carlo using the observed token embeddings:
\begin{align}
\mathrm{KL}(P_{\text{img}}\|P_{\text{txt}}) &\approx 
\frac{1}{N_{\text{img}}}\sum_{i=1}^{N_{\text{img}}} \big(\log P_{\text{img}}(\mathbf{x}_i)-\log P_{\text{txt}}(\mathbf{x}_i)\big), \nonumber\\
\mathrm{KL}(P_{\text{txt}}\|P_{\text{img}}) &\approx
\frac{1}{N_{\text{txt}}}\sum_{i=1}^{N_{\text{txt}}} \big(\log P_{\text{txt}}(\mathbf{y}_i)-\log P_{\text{img}}(\mathbf{y}_i)\big),\nonumber
\end{align}
which is computationally efficient and leverages the empirical embedding sets as observed samples for Monte Carlo approximation.

\subsection{Multi-Scale Aggregation for Caption Evaluation}
\label{sec:fusion}

Most existing caption evaluation methods rely primarily on global semantic similarity between image and text embeddings. Specifically, a global alignment score
\begin{equation}
g(I,T) = \bar{\mathbf{v}}^\top \bar{\mathbf{u}},
\end{equation}
provides a coarse-grained measure of semantic consistency. However, this formulation has a fundamental limitation: it is insensitive to  fine-grained mismatches. Captions that are globally related to the image but contain hallucinated objects, missing attributes, or incorrect relations can still achieve high similarity scores. This indicates that global alignment alone provides a low-resolution but biased estimate of caption quality. To address this issue, we consider token-level or distributional comparisons, leading to a local divergence measure $d(I,T)$ (Eq.~\eqref{eq:weighted_div_refined}). While this signal is highly sensitive to fine-grained inconsistencies, it introduces another challenge: high variance. Token-level comparisons are affected by lexical variation, paraphrasing, and ambiguity in natural language descriptions, which may lead to unstable estimates when used in isolation.

The above analysis reveals a natural trade-off: global similarity provides 
robust but coarse semantic alignment, whereas local divergence provides 
fine-grained but noisy corrective signals. Therefore, these two signals 
are not redundant, but complementary in both semantic granularity and statistical behavior. Motivated by this complementarity, we construct a unified scoring function 
that integrates both signals:
\begin{equation}
\mathrm{MSD}(I,T) = g(I,T) - \alpha \, d(I,T),
\end{equation}
where $\alpha>0$ controls the contribution of the local high-variance signal.

The combination of global alignment and local divergence can be understood  from a risk minimization perspective. These two signals exhibit complementary statistical properties: the global score is relatively stable but biased (insensitive to local inconsistencies), whereas the local divergence is more expressive but noisy due to token-level variability. We model the evaluation of a candidate caption as minimizing a composite risk:
\begin{equation}
\mathcal{R}(T) = \underbrace{\mathbb{E}[-g(I,T)]}_{\text{low-variance semantic prior}} 
+ \underbrace{\lambda \, \mathbb{E}[d(I,T)]}_{\text{high-variance correction}},
\end{equation}
where $\lambda$ corresponding to $\alpha$ balances a bias--variance trade-off. 
In our experiments, we set $\alpha$ at a small value, which is not incidental but reflects the higher variance of local divergence, ensuring that the global semantic prior dominates unless strong contradictory evidence is present.

In multi-candidate settings, the reliability of the global signal varies 
across instances. When global scores are sharply peaked, the ranking is 
confident and additional corrections are likely to introduce noise. 
When scores are flat, the global signal is ambiguous and requires refinement. We model this as a heteroscedastic noise setting, where the variance of the 
correction term depends on the confidence of the global distribution. 
Specifically, we define a softmax over global scores:
\begin{equation}
p_j = \frac{\exp(g_j/\xi)}{\sum_{\ell=1}^{M}\exp(g_\ell/\xi)},
\end{equation}
and measure uncertainty as
\begin{equation}
u =
\begin{cases}
1, & M=1,\\[1mm]
\dfrac{M}{M-1}\left(1-\max_j p_j\right), & M>1.
\end{cases}
\end{equation}
This uncertainty term can be interpreted as a data-dependent estimate of 
the noise level in the global signal. Incorporating this into the risk formulation yields:
\begin{equation}
\mathrm{Soft\mbox{-}MSD}(I,T_j) = g_j - \alpha \,\cdot u \cdot \, d(I,T_j),
\end{equation}
which corresponds to a confidence-aware weighting of the correction term. Importantly, this formulation does not assume that global similarity is 
always reliable. Instead, it uses the \emph{relative separability} of candidates 
as a proxy for confidence: when candidates are indistinguishable under 
global alignment (high uncertainty), local divergence is amplified to 
resolve ambiguity; when one candidate dominates (low uncertainty), the 
method avoids overfitting to high-variance local signals. This perspective also explains robustness to overconfident but incorrect predictions in LLM generated descriptions: such cases typically exhibit large local divergence, and are penalized whenever competing candidates induce non-negligible uncertainty.

\section{Experiments}
\label{sec:experiments}

We evaluate the proposed MSD-Score on fine-grained image--text alignment benchmarks designed to probe localized semantic correctness beyond global relevance. Our experimental setup is organized around three key questions: (i) whether MSD-Score improves reference-free caption evaluation under subtle compositional perturbations, (ii) whether it is effective at detecting fine-grained semantic errors and unsupported content, and (iii) whether the proposed multi-scale design generalizes across diverse vision--language encoders.

\textbf{Datasets.} We evaluate MSD-Score on a diverse set of benchmarks covering human preference alignment, factual correctness, compositional sensitivity, and hallucination detection. Importantly, we distinguish between \emph{human-annotated benchmarks}, which serve as primary evidence, and \emph{controlled diagnostic benchmarks}, which provide complementary stress tests.

\noindent\textbf{Human-annotated benchmarks.} We primarily evaluate on datasets with human judgments, ensuring that conclusions are grounded in real evaluation signals.
\textbf{CapArena}~\cite{cheng2025caparena} focuses on long-form caption quality under a pairwise preference protocol. 
It contains over 6K comparisons where human annotators select the better caption among competing model outputs. 
This benchmark evaluates both caption-level and model-level agreement with human preferences, emphasizing holistic quality and narrative coherence.
\textbf{DocENT (PoSh)}~\cite{ananthram2025posh} targets fine-grained factual correctness and descriptive completeness. 
It provides both pairwise judgments and span-level error annotations, capturing omissions, incorrect attributes, and relational errors. 
The dataset is particularly challenging due to complex scenes (e.g., artwork) and expert-level annotations.

\noindent\textbf{Compositional diagnostic benchmark.}
\textbf{SugarCrepe}~\cite{hsieh2023sugarcrepe} evaluates sensitivity to compositional errors using minimally edited counterfactual captions. 
Each instance consists of a correct caption and a hard negative differing in a single semantic aspect (object, attribute, or relation). 
This benchmark isolates fine-grained semantic reasoning and is widely used to assess robustness against “shortcut” scoring.

\noindent\textbf{Controlled counterfactual diagnostic benchmark (ours).}
\textbf{COCO-CF} is a synthetic diagnostic benchmark constructed in this work to systematically probe hallucination sensitivity. 
It is built from 5,000 images sampled from the MS-COCO validation set \cite{chen2015microsoft}, where each negative caption is created via a \emph{single controlled semantic change} (object, attribute, or relation), forming a pair $(c^+, c^-)$. We emphasize that COCO-CF is not intended to replace human-annotated benchmarks, but rather to serve as a controlled stress test for fine-grained semantic errors. 
All methods are evaluated under a shared pairwise protocol, and hyperparameters are tuned once on a held-out split and fixed across all benchmarks to ensure fairness. To analyze difficulty, COCO-CF defines two splits:
\emph{Easy-CF}, where injected concepts are clearly absent from the image, and 
\emph{Hard-CF}, where confusable or semantically related concepts are introduced. 
Captions are generated using multiple captioning models to ensure diversity\footnote{Valid pairs per captioner (Easy/Hard): LLaVA (5000/4999), InternVL (5000/4997), ChatGPT-4o-mini (4999/4999), and QwenVL (5000/4998).}.

\noindent\textbf{Additional standard benchmarks.}
We further include widely used caption evaluation datasets with human scores for completeness.
\textbf{Flickr8k-Expert}~\cite{hodosh2013framing} provides expert ratings (1–4) for image–caption pairs, enabling correlation-based evaluation.
\textbf{Flickr8k-CF}~\cite{hodosh2013framing} contains crowd-sourced binary relevance judgments, where we follow prior work and use the proportion of positive labels.
\textbf{Composite}~\cite{aditya2015images} aggregates human judgments across Flickr8k \cite{hodosh2013framing}, Flickr30k \cite{young2014image}, and COCO \cite{lin2014microsoft}, with scores ranging from 1 to 5.
\textbf{Pascal-50S}~\cite{vedantam2015cider} evaluates pairwise caption preference across multiple settings (human vs. human, human vs. machine, etc.), with majority voting over 48 annotations per pair.

\textbf{Evaluation Protocol.}
We evaluate caption scoring methods along three complementary dimensions: 
(i) fine-grained discriminative ability, 
(ii) alignment with human preferences, and 
(iii) system-level ranking consistency. 
These dimensions correspond to different failure modes of caption evaluation and are instantiated across multiple benchmarks.

\noindent\textbf{Pairwise discriminative evaluation.}
For datasets constructed with minimal semantic differences (e.g., SugarCrepe and COCO-CF), we adopt a \emph{pairwise preference} protocol. Each instance consists of an image $I$ and a caption pair $(c^+, c^-)$, where $c^+$ is semantically aligned and $c^-$ contains a controlled error. Given a scoring function $S(I,c)$, we measure \textbf{pairwise accuracy}:
\begin{equation}
\mathrm{Acc}
=
\frac{1}{N}
\sum_{i=1}^{N}
\mathbb{I}\!\left[
S(I_i, c_i^+) > S(I_i, c_i^-)
\right],
\end{equation}
which directly evaluates sensitivity to fine-grained semantic discrepancies 
while controlling for linguistic fluency.

\noindent\textbf{Caption-level human agreement.}
On human-annotated benchmarks such as CapArena, we measure how well a metric aligns with human preferences over caption pairs. Given a caption pair $(c_1, c_2)$ and human label $y \in \{>, <, =\}$, we derive the predicted preference $\hat{y}$ from the score difference $\Delta S = S(I,c_1)-S(I,c_2)$: a tie is predicted if $|\Delta S|\le \epsilon_{\text{tie}}$, and otherwise the higher-scoring caption is selected. 
We set $\epsilon_{\text{tie}}=10^{-4}$ for all methods. \textbf{Caption-level agreement} is defined as the proportion of instances where $\hat{y}=y$. 
This metric emphasizes robustness in scenarios where multiple captions may be globally plausible.

\noindent\textbf{System-level ranking consistency.}
Beyond pairwise decisions, we evaluate whether a metric induces consistent rankings over captioning models. Following CapArena, we compute rank correlations between metric-induced model rankings and human rankings,  including Spearman’s $\rho$ and Kendall’s $\tau$:
\begin{equation}
\rho = 1 - \frac{6\sum_i d_i^2}{M(M^2-1)}, 
\qquad
\tau = \frac{n_c - n_d}{\tbinom{M}{2}},
\end{equation}
where $d_i$ is the rank difference for model $i$, and $n_c$, $n_d$ denote the number of concordant and discordant model pairs.

\noindent\textbf{Fine-grained error sensitivity.}
On DocENT (PoSh), we further evaluate sensitivity to specific error types, including factual mistakes and omissions. We report both pairwise accuracy and rank correlation with human ratings, as well as overall quality agreement. This setup jointly assesses local error detection and global descriptive fidelity.

Together, these protocols provide a comprehensive evaluation: pairwise accuracy measures fine-grained discrimination, caption-level agreement captures alignment with human judgment, and rank correlation evaluates consistency at the system level.

\begin{table*}[!t]
  \centering
  \renewcommand{\arraystretch}{1.2}
  \caption{\textbf{Alignment with human preferences on CapArena.} Soft-MSD achieves the strongest caption-level agreement among reference-free metrics and matches or exceeds reference-based metrics, while remaining fully reference-free. In controlled 7B comparisons, Soft-MSD is competitive with VLM-as-a-Judge at the caption level, highlighting its effectiveness without relying on generative judges.
}
  \label{tab:caparena}
  \resizebox{\textwidth}{!}{
  \begin{tabular}{l|c|ccccc|cc}
    \thickhline
    \multirow{2}{*}{\textbf{Metric}} & \textbf{Need}
    & \multicolumn{5}{c|}{\textbf{Caption-level Agreement (Including Ties)}}
    & \multicolumn{2}{c}{\textbf{Model-level Agreement}} \\
     & \textbf{Ref?}
     & \textbf{Overall}
     & \textbf{Level1}
     & \textbf{Level2}
     & \textbf{Level3}
     & \textbf{Level4}
     & \textbf{Spearman $\rho$}
     & \textbf{Kendall $\tau$} \\
    \hline
    BLEU-4 \cite{papineni2002bleu} & Yes & 47.4 & 47.7 & 48.0 & 47.5 & 46.7 & 0.424 & 0.319 \\
    SPICE \cite{anderson2016spice} & Yes & 41.7 & 44.1 & 42.2 & 41.5 & 38.7 & 0.275 & 0.231 \\
    CIDER \cite{vedantam2015cider} & Yes & 38.4 & 37.8 & 38.3 & 38.9 & 38.7 & $-0.279$ & $-0.209$ \\
    Polos \cite{wada2024polos} & Yes & 47.9 & 52.6 & 46.7 & 46.2 & 46.2 & 0.420 & 0.363 \\
    CAPTURE \cite{dong2024benchmarking} & Yes & 52.5 & 60.1 & 51.2 & 50.4 & 47.9 & 0.613 & 0.538 \\
    VDC-Score \cite{chai2024auroracap} & Yes & 55.7 & \textbf{68.7} & 57.9 & 49.6 & 46.0 & \textbf{0.890} & \textbf{0.736} \\
    METEOR \cite{banerjee2005meteor} & Yes & \textbf{57.6} & 65.7 & \textbf{58.2} & \textbf{53.6} & \textbf{53.0} & 0.785 & 0.582 \\
    \hline
    CLIPScore \cite{hessel2021clipscore} & No & 32.5 & 26.6 & 30.8 & 36.2 & 35.5 & $-0.574$ & $-0.451$ \\
    LongCLIPScore \cite{zhang2024long} & No & 40.0 & 42.2 & 40.4 & 38.4 & 39.5 & $-0.226$ & $-0.121$ \\
    FLEUR \cite{lee2024fleur} & No & 45.8 & 51.3 & 46.2 & 44.4 & 41.4 & 0.393 & 0.297 \\
    \textbf{Soft-MSD (ours)} & No & \textbf{57.6} & \textbf{64.4} & \textbf{58.7} & \textbf{54.2} & \textbf{53.6} & \textbf{0.398} & \textbf{0.297} \\
    \thickhline
    \multicolumn{9}{l}{\textit{Controlled comparisons with 7B model for reference-free comparison}} \\
    \hline
    VLM-as-a-Judge (LLaVA-7B~\cite{liu2023visual-llava}) & No & 46.1 & 49.4 & 45.4 & 45.0 & 44.6 & 0.288 & 0.187 \\
    Soft-MSD (LLaVA-7B~\cite{liu2023visual-llava}) & No & \textbf{57.6} & \textbf{64.4} & \textbf{58.7} & \textbf{54.2} & \textbf{53.6} & \textbf{0.398} & \textbf{0.297} \\
    \thickhline
    VLM-as-a-Judge (LLaVA-OneVision-7B~\cite{li2024llava}) & No & 58.8 & \textbf{67.0} & 61.5 & 55.8 & 51.1 & \textbf{0.698} & 0.533 \\
    Soft-MSD (LLaVA-OneVision-7B~\cite{li2024llava}) & No & \textbf{59.2} & 66.5 & \textbf{61.7} & \textbf{58.5} & \textbf{52.8} & 0.692 & \textbf{0.565} \\
    \thickhline
    VLM-as-a-Judge (LLaVA-Critic-7B~\cite{xiong2025llava}) & No & 53.5 & \textbf{66.7} & 52.9 & 48.1 & 46.3 & \textbf{0.888} & \textbf{0.693} \\
    Soft-MSD (LLaVA-Critic-7B~\cite{xiong2025llava}) & No & \textbf{55.2} & 56.0 & \textbf{56.3} & \textbf{53.3} & \textbf{55.7} & 0.511 & 0.440 \\
    \thickhline
    VLM-as-a-Judge (Qwen2.5-VL-7B~\cite{bai2025qwen2}) & No & 52.2 & 59.8 & 52.1 & 49.3 & 47.3 & \textbf{0.903} & \textbf{0.780} \\
    Soft-MSD (Qwen2.5-VL-7B~\cite{bai2025qwen2}) & No & \textbf{56.7} & \textbf{62.6} & \textbf{57.8} & \textbf{53.1} & \textbf{54.9} & 0.832 & 0.726 \\
    \thickhline
  \end{tabular}}
  \vspace{-4mm}
\end{table*}

\textbf{Implementation Details.}
We first summarize the overall pipeline, followed by training settings, hyperparameters, and efficiency. Our method builds on a vision--language encoder to extract patch-level and token-level embeddings, followed by distributional modeling via von Mises--Fisher (vMF) mixtures. Local divergence is computed from these distributions and combined with global similarity to produce MSD and Soft-MSD scores. Importantly, all inference-time computations are lightweight and do not require LLMs.

\noindent\textbf{Backbone and local alignment.}
Unless otherwise specified, we use CLIP ViT-L/14 as the visual backbone. 
A lightweight local vision--language alignment model is trained on CC3M-595K 
following the LLaVA data construction protocol. Image features are projected into the language embedding space via a linear layer. A frozen LLaMA-7B is used only during training for reconstruction supervision and is \emph{not} invoked during inference.

\noindent\textbf{Distribution modeling.}
Both image patches and text tokens are modeled using vMF mixtures with fixed concentration $\kappa=20$. Mixture parameters are estimated using EM with a fixed iteration budget ($m=20$). We set mixture sizes based on caption length regimes: 
$(K_{\mathrm{img}},K_{\mathrm{txt}})=(3,2)$ for short captions (SugarCrepe, COCO-CF), 
and $(5,3)$ for long captions (CapArena, DocENT). These choices are fixed \emph{a priori} and not tuned per dataset.

\noindent\textbf{Hyperparameters.}
All scalar hyperparameters are tuned once on a held-out validation split (1K samples from SugarCrepe) and kept fixed across all benchmarks. This includes the fusion weight $\alpha=0.1$, Soft-MSD temperature $\xi=0.2$, and length-adaptive Bi-KL parameters $(L_0=20, \tau_L=3.0)$. This protocol ensures fair comparison without dataset-specific tuning.

\noindent\textbf{EM details.}
We initialize mixture components by sampling tokens or patches and normalizing them to unit length, with uniform mixture weights. During EM, components with negligible effective counts ($<10^{-6}$) are reinitialized to avoid collapse. All runs use a fixed number of iterations for stability.

\noindent\textbf{Baselines and auxiliary methods.}
We include a simple rank-aggregation baseline (RankAgg) for analysis:
\[
\hat{y}=
\begin{cases}
\mathbb{I}(S_{\text{cos}}^{+} > S_{\text{cos}}^{-}), & \text{if } |S_{\text{cos}}^{+}-S_{\text{cos}}^{-}| > \tau_r,\\
\mathbb{I}(D_{\text{Bi-KL}}^{+} < D_{\text{Bi-KL}}^{-}), & \text{otherwise.}
\end{cases}
\]
RankAgg is used only as a diagnostic baseline (Figure~\ref{fig:results1}) and is not part of the proposed method.

\noindent\textbf{VLM-as-a-Judge protocol.}
For controlled comparisons with LLM-based evaluators (Table~\ref{tab:caparena}), 
we follow the CapArena VLM-as-a-Judge setup. The judge receives an image and two candidate captions, and produces a preference judgment based on multiple criteria (e.g., precision, informativeness, hallucination). We use the reference-free setting with open-source 7B models. Decoding parameters are fixed across all runs (\texttt{top-p=0.9}, \texttt{temperature=0.5}).

\subsection{Comparison with State-of-the-art Methods}

\begin{table*}[!t]
  \centering
  \renewcommand{\arraystretch}{1.25}
  \caption{\textbf{Sensitivity to factual errors and omissions on DocENT (PoSh).} Soft-MSD consistently outperforms prior reference-free metrics across mistakes, omissions, and overall quality, demonstrating strong fine-grained error detection capability.
}
  \label{tab:docent}
  \resizebox{0.8\textwidth}{!}{
  \begin{tabular}{l|ccc|ccc|ccc}
    \thickhline
    \multirow{2}{*}{\textbf{Metric}} &
    \multicolumn{3}{c|}{\textbf{Mistakes}} &
    \multicolumn{3}{c|}{\textbf{Omissions}} &
    \multicolumn{3}{c}{\textbf{Overall Quality}} \\
    \cline{2-10}
    & Acc & $\rho$ & $\tau$ & Acc & $\rho$ & $\tau$ & Acc & $\rho$ & $\tau$ \\
    \hline
    CLIPScore \cite{hessel2021clipscore} & 45.3 & 0.145 & 0.108 & 47.0 & 0.176 & 0.133 & 53.5 & 0.181 & 0.136 \\
    FLEUR \cite{lee2024fleur} & 35.2 & $-0.053$ & $-0.040$ & 38.5 & 0.029 & 0.020 & 41.2 & $-0.040$ & $-0.031$ \\
    \textbf{Soft-MSD (ours)} & \textbf{64.4} & \textbf{0.219} & \textbf{0.165} & \textbf{62.1} & \textbf{0.219} & \textbf{0.164} & \textbf{64.8} & \textbf{0.259} & \textbf{0.195} \\
    \thickhline
  \end{tabular}}
\end{table*}

\begin{table*}[!t]
  \centering
  \renewcommand{\arraystretch}{1.15}
  \caption{ \textbf{Comparison on legacy benchmarks for broader positioning.} MSD-Score achieves competitive or state-of-the-art performance across standard benchmarks,  demonstrating strong generalization beyond specialized evaluation settings.
}
  \label{tab:prior_comparison}
  \setlength{\tabcolsep}{5pt}
  \begin{tabular}{l|cc|cc|cc|ccccc}
    \thickhline
    \multirow{2}{*}{\textbf{Metric}} & \multicolumn{2}{c|}{\textbf{Flickr8k-Expert} \cite{hodosh2013framing}} & \multicolumn{2}{c|}{\textbf{Flickr8k-CF} \cite{hodosh2013framing}} & \multicolumn{2}{c|}{\textbf{Composite} \cite{aditya2015images}} & \multicolumn{5}{c}{\textbf{Pascal-50S} \cite{vedantam2015cider}} \\
    & $\tau_b$ & $\tau_c$ & $\tau_b$ & $\tau_c$ & $\tau_b$ & $\tau_c$ & HC & HI & HM & MM & Mean \\
    \hline
    UMIC \cite{lee2021umic} & - & 46.8 & - & - & - & 56.1 & 66.1 & \textbf{99.8} & \textbf{98.1} & 76.2 & 85.1 \\
    PAC-S \cite{sarto2023pacs} & 53.9 & 54.3 & 36.0 & 18.6 & 51.5 & 55.7 & 60.6 & 99.3 & 96.9 & 72.9 & 82.4 \\
    BRIDGE (ViT-B/32) \cite{sarto2024bridge} & 54.4 & 54.8 & 36.1 & 18.7 & 50.9 & 55.0 & 59.4 & 99.4 & 97.5 & 74.0 & 82.6 \\
    BRIDGE (ViT-L/14) \cite{sarto2024bridge} & 55.4 & 55.8 & 36.3 & 19.0 & 52.9 & 57.2 & 61.2 & 99.6 & 96.6 & 74.1 & 82.9 \\
    InfoMeTIC \cite{hu2023infometic} & - & 54.2 & 36.3 & - & - & 59.2 & 69.0 & \textbf{99.8} & 94.0 & 78.3 & 85.3 \\
    InfoMeTIC+ \cite{hu2023infometic} & - & 55.5 & 36.6 & - & - & 59.3 & 69.9 & 99.7 & 96.8 & 79.6 & 86.5 \\
    HICE-S \cite{zeng2024hicescore} & 55.9 & 56.4 & 37.2 & 19.2 & 53.1 & 57.9 & 68.6 & 99.7 & 96.5 & 79.5 & 86.1 \\
    EXPERT \cite{kim2025expert} & - & 56.7 & 39.3 & - & - & \textbf{65.0} & 62.8 & \text{99.8} & 97.8 & 78.4 & 84.7 \\
    \textbf{Soft-MSD (ours)} & \textbf{57.5} & \textbf{58.3} & \textbf{39.4} & \textbf{19.7} & \textbf{60.3} & 64.2 & \textbf{69.4} & 99.7 & 97.1 & \textbf{81.2} & \textbf{86.9} \\
    \thickhline
  \end{tabular}
  \vspace{-4mm}
\end{table*}

\textbf{Alignment with human preferences (CapArena).}
We first evaluate alignment with human judgments on CapArena, a challenging benchmark for long-form caption quality. As shown in Table~\ref{tab:caparena}, Soft-MSD achieves the strongest caption-level agreement among all reference-free metrics, and matches or exceeds several reference-based metrics despite not relying on ground-truth captions. This result highlights a key property of MSD-Score: by combining global semantic alignment with fine-grained distributional verification, it captures both coarse relevance and local correctness, leading to improved agreement with human preferences. At the model level, rank correlations remain lower than the strongest reference-based metrics. This suggests that system-level ranking of captioning models is inherently more challenging than pairwise caption discrimination, and may require additional calibration beyond local consistency signals.

\textbf{Comparison with VLM-as-a-Judge.} The lower block of Table~\ref{tab:caparena} presents a controlled comparison with VLM-as-a-Judge under a unified 7B setting. Across multiple backbones, Soft-MSD consistently achieves competitive or superior caption-level agreement, despite being a deterministic and lightweight metric. This result is notable because judge-based methods rely on large generative models, whereas MSD-Score operates without inference-time LLMs. The findings suggest that fine-grained distributional verification provides a strong complementary signal to generative evaluation.

\textbf{Fine-grained error sensitivity (DocENT).}
We next evaluate sensitivity to factual errors and omissions on DocENT. 
As shown in Table~\ref{tab:docent}, Soft-MSD substantially outperforms prior reference-free metrics across all criteria, achieving large gains in both accuracy and correlation with human judgments. These improvements indicate that local distributional modeling effectively captures semantic discrepancies that are missed by holistic similarity metrics. In particular, the consistent gains across mistakes and omissions 
support the claim that MSD-Score improves both error detection and completeness evaluation.

\textbf{Broader comparison on legacy benchmarks.}
Finally, Table~\ref{tab:prior_comparison} reports results on widely used legacy benchmarks. MSD-Score achieves competitive or state-of-the-art performance across these datasets, demonstrating strong generalization beyond specialized evaluation settings. We note that these benchmarks are less sensitive to fine-grained semantic errors, 
and therefore serve primarily as complementary evidence. Our main conclusions are drawn from CapArena and DocENT, which provide more direct evaluation of human alignment and error sensitivity.

\begin{figure}[!t]
    \centering
    \includegraphics[width=\columnwidth]{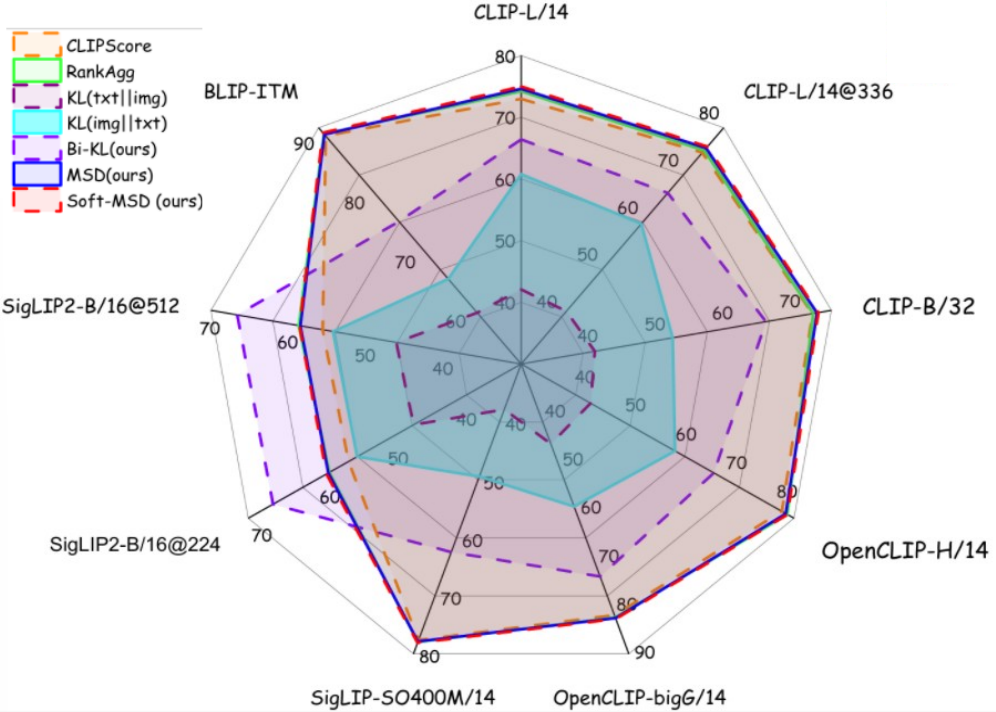}
    \vspace{-4mm}
    \caption{SugarCrepe evaluation on fine-grained compositional errors in short captions. We report pairwise accuracy on minimal counterfactual edits involving objects and attributes.}
    \label{fig:results1}
    \vspace{-4mm}
\end{figure}

\begin{figure}[!t]
    \centering
    \includegraphics[width=0.95\columnwidth]{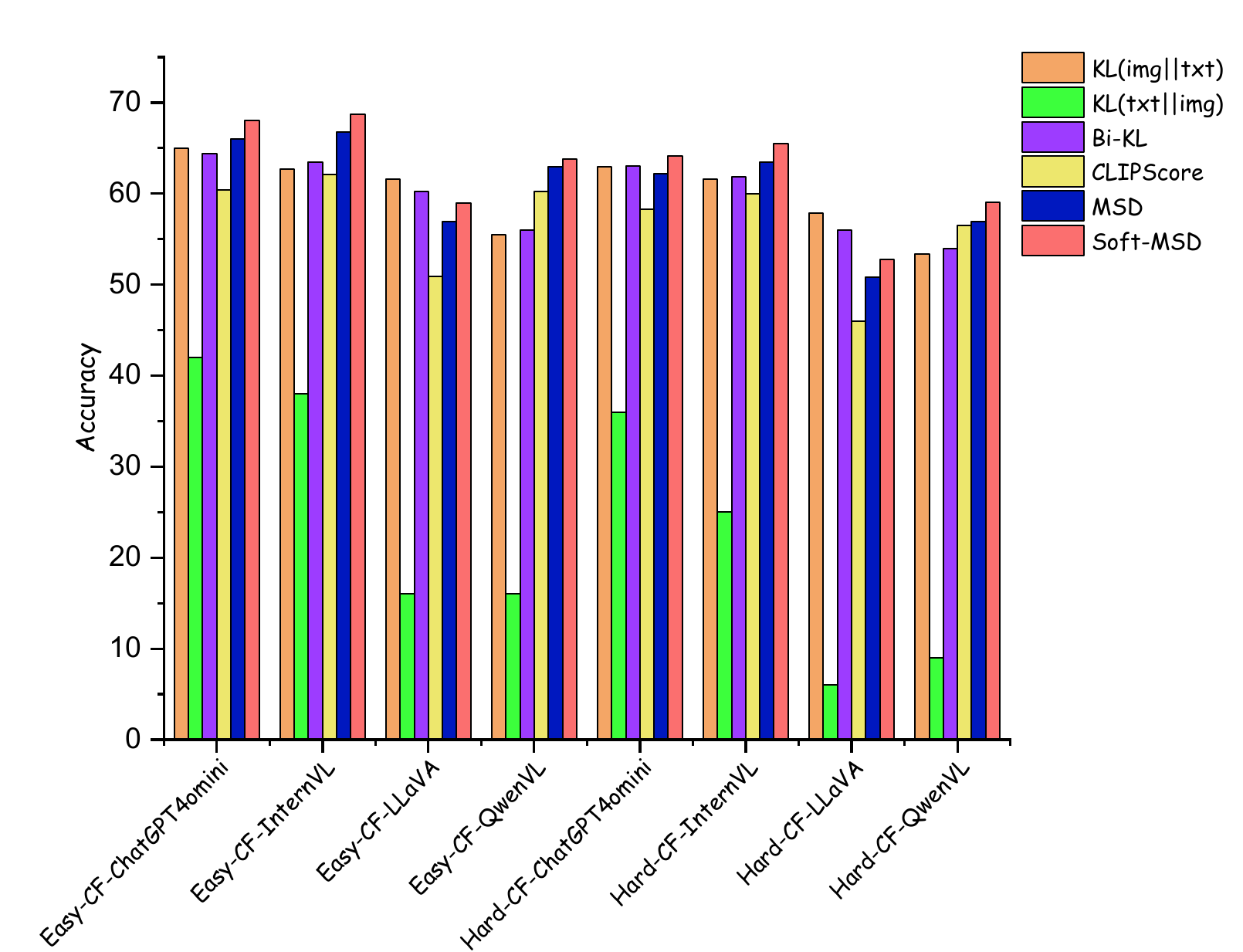}
    \vspace{-4mm}
    \caption{COCO-CF performance by caption source. Pairwise accuracy on Easy-CF and Hard-CF across different caption generators. The bar chart separates caption sources more clearly and highlights the consistent gains of MSD/Soft-MSD on the harder counterfactual settings.}
    \label{fig:results2}
    \vspace{-4mm}
\end{figure}

\textbf{Controlled compositional and hallucination evaluation.}
On SugarCrepe and COCO-CF, which isolate minimal semantic differences, MSD and Soft-MSD consistently improve pairwise discrimination accuracy across backbones and caption sources (Fig.~\ref{fig:results1}, Fig.~\ref{fig:results2}, Table~\ref{tab:cococf_overall}). These results provide controlled evidence that the proposed local divergence term enhances sensitivity to fine-grained semantic changes, including hallucinated objects, incorrect attributes, and relational inconsistencies.

\begin{table*}[!t]
  \centering
  \renewcommand{\arraystretch}{1.25}
   \caption{COCO-CF: Pairwise accuracy (\%) on the controlled diagnostic benchmark for counterfactual hallucination detection. Results are \textbf{micro-averaged} over all valid (image, captioner) pairs after filtering ($N_{\text{easy}}=19{,}999$, $N_{\text{hard}}=19{,}993$). Incorporating local divergence substantially improves robustness to unsupported content, while multi-scale fusion yields the best overall performance.}
  \label{tab:cococf_overall}
  \resizebox{0.8\linewidth}{!}{
  \begin{tabular}{l|cccccc}
    \thickhline
    \textbf{Split} & $\mathrm{KL}\!\left(P_{\text{img}} \,\|\, P_{\text{txt}}\right)$ & $\mathrm{KL}\!\left(P_{\text{txt}} \,\|\, P_{\text{img}}\right)$ & \textbf{Bi-KL} & \textbf{CLIPScore} & \textbf{MSD} & \textbf{Soft-MSD (ours)} \\
    \hline
    Easy-CF & 61.20 & 28.00 & 61.02 & 59.19 & \textbf{63.93} & 63.89 \\
    Hard-CF & 58.97 & 19.00 & 58.71 & 56.20 & 59.34 & \textbf{59.38} \\
    \thickhline
  \end{tabular}}
  \vspace{-4mm}
\end{table*}

\subsection{Mechanism Validation, Efficiency, and Generalization}

\textbf{Component contributions (multi-scale validation).}
We first analyze the contribution of individual components to validate the proposed multi-scale formulation. As shown in Fig.~\ref{fig:results1}, Fig.~\ref{fig:results2}, and Table~\ref{tab:cococf_overall}, local divergence terms consistently improve discrimination on hard compositional and counterfactual cases, where global similarity alone is insufficient. The fused variants (MSD and Soft-MSD) achieve the most stable performance across settings, confirming that global alignment provides a robust prior while local divergence supplies fine-grained corrective signals. Soft-MSD further improves stability by adaptively down-weighting noisy local signals under low uncertainty, supporting the effectiveness of the proposed uncertainty-aware fusion.

\textbf{Robustness across backbones and caption sources.}
We evaluate robustness across diverse vision--language backbones and caption generators. Across SugarCrepe and COCO-CF (Fig.~\ref{fig:results1} and Fig.~\ref{fig:results2}), MSD and Soft-MSD exhibit consistently strong performance with minimal variance across models. A minor deviation is observed for LLaVA on COCO-CF, where the image$\rightarrow$text KL term slightly outperforms. We attribute this to model-specific training biases. Importantly, the fused variants remain competitive in this case and outperform single components in the majority of settings, demonstrating that the multi-scale formulation generalizes robustly across architectures.

\textbf{Efficiency and scalability.}
We profile per-(image, caption) inference time on an RTX A6000 over 21,974 pairs. 
As shown in Table~\ref{tab:latency_breakdown}, CLIPScore requires 34.98 ms per pair, 
while MSD and Soft-MSD require 43.39 ms and 43.44 ms, respectively. Runtime is dominated by feature extraction (34.86 ms), while the additional EM and divergence computation introduces only a modest overhead (8.39 ms). Importantly, this overhead remains nearly constant across caption lengths (8.23 / 8.39 / 8.57 ms), indicating good scalability. This behavior aligns with our design: local token sets and mixture sizes are small, and EM is run with a fixed iteration budget. Under fixed mixture size and iteration count, both EM fitting and KL computation scale linearly with the number of tokens, resulting in mild computational overhead in practice. Overall, MSD-Score introduces a small constant-time overhead relative to embedding extraction, 
making it suitable for high-fidelity offline evaluation.

\begin{table}[!t]
  \centering
  \renewcommand{\arraystretch}{1.15}
  \caption{Per-(image, caption) inference timing (ms). Feature extraction dominates runtime, while EM/KL adds a modest and nearly length-stable overhead.}
  \label{tab:latency_breakdown}
  \resizebox{0.48\textwidth}{!}{
  \begin{tabular}{l|r|ccccc}
    \thickhline
    \textbf{Caption length bucket} & \textbf{Pairs} & \textbf{Feature} & \textbf{EM/KL} & \textbf{CLIPScore} & \textbf{MSD} & \textbf{Soft-MSD} \\
    \hline
    $\leq 16$ tokens & 7{,}843 & 28.22 & 8.23 & 28.34 & 36.59 & 36.63 \\
    17--126 tokens & 6{,}834 & 35.33 & 8.39 & 35.44 & 43.86 & 43.91 \\
    $>126$ tokens & 7{,}297 & 41.54 & 8.57 & 41.68 & 50.26 & 50.31 \\
    \hline
    \textbf{Overall} & \textbf{21{,}974} & \textbf{34.86} & \textbf{8.39} & \textbf{34.98} & \textbf{43.39} & \textbf{43.44} \\
    \thickhline
  \end{tabular}}
\end{table}

\textbf{Generalization under domain shift.}
We further evaluate generalization under domain shift using two out-of-domain benchmarks: ROCOv2~\cite{ruckert2024rocov2} (medical) and RSICD~\cite{lu2018exploring} (remote sensing). Without additional alignment training, we directly replace the backbone with domain-specific encoders 
(BiomedCLIP~\cite{zhang2023biomedclip} and RemoteCLIP~\cite{liu2023remoteclip}). As shown in Table~\ref{tab:ood_transfer}, MSD and Soft-MSD consistently outperform CLIPScore, achieving 2--3\% absolute gains across both domains. 
These results indicate that the proposed multi-scale scoring formulation generalizes beyond the original training distribution. More broadly, MSD-Score should be viewed as a \emph{domain-adaptable evaluation layer}: the scoring formulation (vMF + bi-directional KL) is model-agnostic and remains unchanged, while only the underlying encoder may require domain-specific adaptation. This decoupled design enables reuse across datasets and tasks within a domain with minimal additional cost.

\begin{table}[!t]
  \centering
  \renewcommand{\arraystretch}{1.15}
  \caption{OOD transfer with domain-specific encoders. We report pairwise accuracy on two out-of-domain benchmarks without additional alignment training.}
  \label{tab:ood_transfer}
  \begin{tabular}{l|l|ccc}
    \thickhline
    \textbf{Dataset} & \textbf{Backbone} & \textbf{CLIPScore} & \textbf{MSD} & \textbf{Soft-MSD} \\
    \hline
     ROCOv2~\cite{ruckert2024rocov2} & BiomedCLIP~\cite{zhang2023biomedclip} & 71.7 & 73.3 &\textbf{74.9} \\
    RSICD~\cite{lu2018exploring} & RemoteCLIP~\cite{liu2023remoteclip} & 77.6 & 79.0 & \textbf{80.5} \\
    \thickhline
  \end{tabular}
\end{table}

\subsection{Ablation Study}

\textbf{Ablation on concentration $\kappa$: stability--expressivity trade-off.}
We first study the role of the concentration parameter $\kappa$ in vMF modeling, which controls the trade-off between distributional sharpness and robustness. A small $\kappa$ leads to overly diffuse distributions that blur fine-grained semantic distinctions, while an excessively large $\kappa$ results in near-deterministic assignments, making EM updates sensitive to noise and initialization instability. Therefore, the key question is not only sensitivity, but whether a stable intermediate regime exists for reliable local semantic modeling. Table~\ref{tab:kappa_sweep} shows that performance remains stable across a wide range of $\kappa$ values, indicating that the model is not overly sensitive to concentration tuning. Among them, $\kappa=20$ achieves the best balance between discrimination and stability, and is therefore adopted in all experiments.

\begin{table}[!t]
  \centering
  \renewcommand{\arraystretch}{1.25}
  \caption{Effect of vMF concentration $\kappa$ on fine-grained compositional discrimination (MSD-Score). Stability across $\kappa$ indicates robustness of hyperspherical local modeling.}
  \label{tab:kappa_sweep}
  \begin{tabular}{l|ccc}
    \thickhline
    \textbf{Subtask} 
    & $\boldsymbol{\kappa=10}$ 
    & $\boldsymbol{\kappa=20}$ 
    & $\boldsymbol{\kappa=40}$ \\
    \hline
    Add-Att     & 71.53 & \textbf{76.88} & 74.42 \\
    Add-Obj     & 78.18 & \textbf{79.39} & 78.81 \\
    Replace-Att & 78.93 & \textbf{80.33} & 80.08 \\
    Replace-Obj & \textbf{94.07} & \textbf{94.07} & 93.89 \\
    Replace-Rel & 64.30 & 63.44 &\textbf{66.36} \\
    Swap-Att    & 63.06 & \textbf{64.26} & 62.01 \\
    Swap-Obj    & 60.82 & \textbf{63.67} & 60.41 \\
    \thickhline
  \end{tabular}
  \vspace{-4mm}
\end{table}

\begin{figure}[!t]
  \centering
  \includegraphics[width=\linewidth]{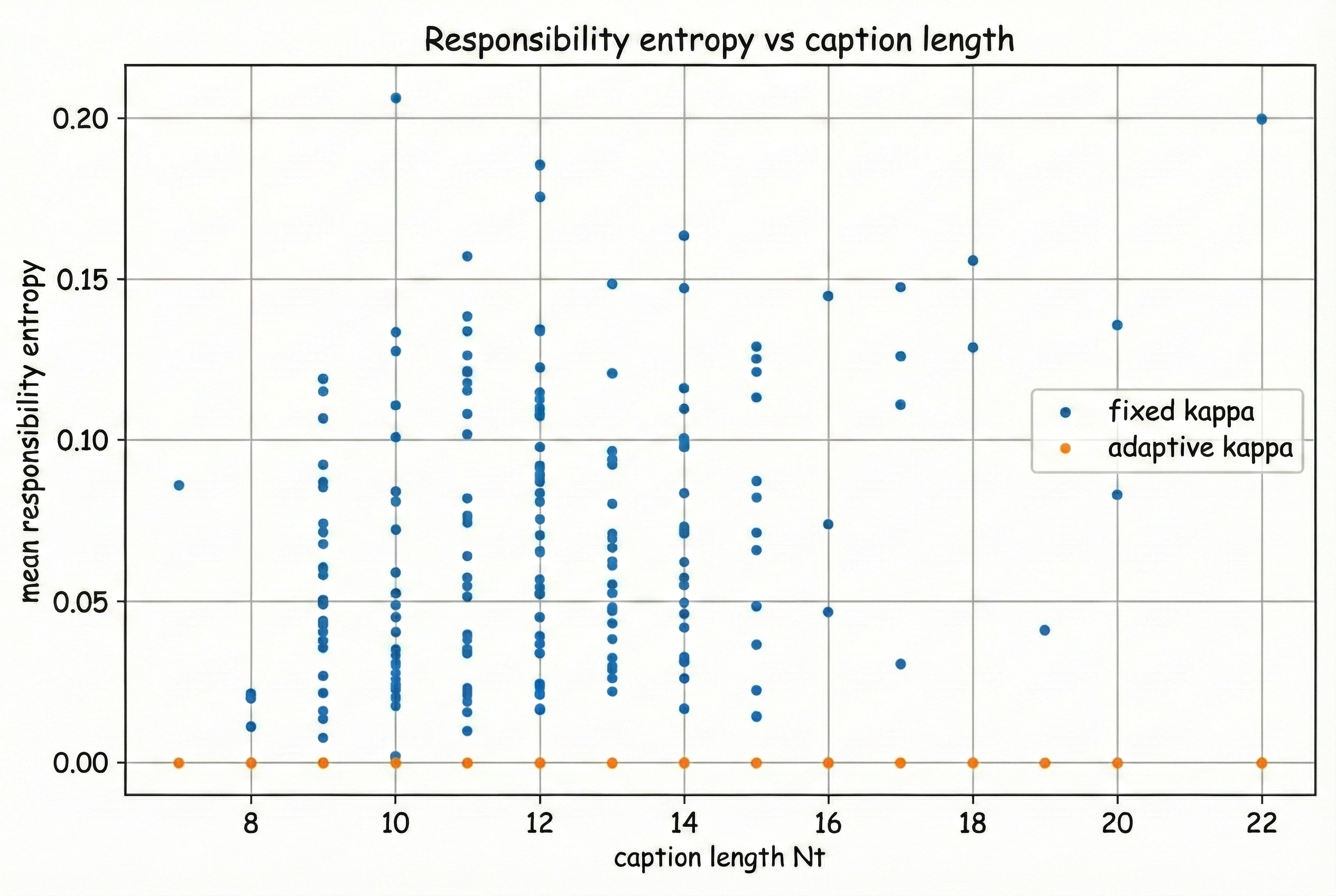}
  \vspace{-4mm}
  \caption{Collapse of EM responsibility entropy under adaptive $\kappa$ vs.\ stability of fixed $\kappa$. Adaptive concentration leads to near-hard assignments in sparse-token regimes, while fixed $\kappa$ preserves uncertainty for stable KL estimation.}
  \label{fig:resp_entropy_vs_length}
  \vspace{-4mm}
\end{figure}

\textbf{Why fixed $\kappa$ is more stable than adaptive $\kappa$ in few-token regimes.}
Although $\kappa$ can be adapted per component, we find that such adaptivity is unstable in the few-token regime typical of image--caption evaluation. Let $r_{nk}=p(z=k \mid x_n)$ denote EM responsibilities. We measure their entropy:
\[
H(x_n) = -\sum_{k=1}^{K} r_{nk}\log r_{nk}.
\]
As shown in Figure~\ref{fig:resp_entropy_vs_length}, adaptive $\kappa$ frequently collapses to near-hard assignments ($H \approx 0$), especially when token counts are small. This removes uncertainty information that is essential for stable KL estimation and leads to high variance in local divergence. In contrast, fixed $\kappa$ preserves soft assignments across caption lengths, yielding smoother aggregation and more stable semantic mismatch signals.

% \textbf{Spherical geometry of vision--language embeddings.}
% We further justify hyperspherical modeling by analyzing embedding norms. As shown in Figure~\ref{fig:embedding_norms}, raw patch and token embeddings exhibit large norm variance, while $\ell_2$ normalization collapses them tightly around $\|x\|_2 \approx 1$. This indicates that vision--language embeddings naturally lie on a hypersphere, making von Mises--Fisher modeling more appropriate than Euclidean Gaussian assumptions.

% \begin{figure*}[!t]
%   \centering
%   \includegraphics[width=0.95\textwidth]{picture/norm_hist_beauty_no_watermark.png}
%   \caption{Raw vs.\ normalized embedding norms. Raw embeddings exhibit large variance, while normalized embeddings concentrate on the unit hypersphere, motivating vMF-based modeling.}
%   \label{fig:embedding_norms}
% \end{figure*}

\textbf{Regime-dependent KL direction: from variance to hallucination sensitivity.}
A key design choice in MSD is the KL direction, which we find to be regime-dependent.
\textbf{Short captions (variance-dominated regime).}
When captions are short, the limited number of text tokens leads to high variance in estimating $\mathrm{KL}(P_{\text{txt}} \| P_{\text{img}})$. In this regime, image patches provide a denser sampling space, making $\mathrm{KL}(P_{\text{img}} \| P_{\text{txt}})$ more stable.
\textbf{Long captions (hallucination-dominated regime).}
Table \ref{tab:kl_length_bucket} breaks down CapArena and SugarCrepe pairs into caption-length quantiles and confirms this regime--dependent behavior. As captions grow longer, token coverage becomes sufficient, and the dominant error shifts to hallucination. In this case, image-averaged KL may dilute localized errors, while text-averaged KL explicitly penalizes unsupported tokens. Thus, $\mathrm{KL}(P_{\text{txt}} \| P_{\text{img}})$ becomes more effective. Bi-KL integrates both perspectives, ensuring both coverage (no missing visual evidence) and faithfulness (no hallucinated content), leading to stable performance across regimes. The results in Table \ref{tab:kl_direction}  empirically corroborate this regime--dependent behavior.

\begin{table*}[!t]
  \centering
  \renewcommand{\arraystretch}{1.15}
  \caption{Effect of KL direction across caption lengths.
 We randomly sampled pairs from CapArena and Sugarcrepe and partitioned them into 6 buckets based on caption token length and report the accuracy (\%) of different KL variants. The preferred direction shifts from $\mathrm{KL}(P_{\text{img}}\|P_{\text{txt}})$ in short captions to $\mathrm{KL}(P_{\text{txt}}\|P_{\text{img}})$ in long captions, while Bi-KL remains robust across buckets.}
  \label{tab:kl_length_bucket}
  \setlength{\tabcolsep}{7pt}
  \begin{tabular}{c|ccc|ccc}
    \thickhline
    \multirow{2}{*}{\textbf{Bucket}} & \multicolumn{3}{c|}{\textbf{Statistics}} & \multicolumn{3}{c}{\textbf{Accuracy (\%)}} \\
     & \textbf{Len Range} & \textbf{Pairs} & \textbf{Mean Len} & $\mathrm{KL}(P_{\text{img}}\|P_{\text{txt}})$ & $\mathrm{KL}(P_{\text{txt}}\|P_{\text{img}})$ & \textbf{Bi-KL} \\
    \hline
    $B_1$ & $[0,13)$     & 1677 & 11.4  & 66.5 & 38.7 & \textbf{79.9} \\
    $B_2$ & $[13,15)$    & 1783 & 13.5  & 64.2 & 39.4 & \textbf{81.4} \\
    $B_3$ & $[15,86)$    & 2021 & 24.2  & 61.4 & 44.2 & \textbf{73.9} \\
    $B_4$ & $[86,147)$   & 1844 & 119.5 & 48.1 & 57.6 & \textbf{58.5} \\
    $B_5$ & $[147,227)$  & 1822 & 183.1 & 54.8 & 72.8 & \textbf{73.9} \\
    $B_6$ & $[227,\infty)$ & 1845 & 370.7 & 67.4 & \textbf{88.0} & \textbf{88.0} \\
    \thickhline
  \end{tabular}
\end{table*}

\begin{table*}[!t]
  \centering
  \renewcommand{\arraystretch}{1.25}
  \caption{Regime-dependent effect of KL direction in long-caption evaluation (CapArena). The optimal direction shifts from image-dominant to text-sensitive regimes, while Bi-KL remains stable.}
  \label{tab:kl_direction}
  \setlength{\tabcolsep}{10pt}
   \begin{tabular}{l|ccccc|cc}
    \thickhline
    \multirow{2}{*}{\textbf{Metric}} & \multicolumn{5}{c|}{\textbf{Caption-level Agreement (Including Tie)}} & \multicolumn{2}{c}{\textbf{Model-level Agreement}} \\
     & \textbf{Overall} & \textbf{Level1} & \textbf{Level2} & \textbf{Level3} & \textbf{Level4} & \textbf{Spearman} & \textbf{Kendall} \\
    \hline
    $\text{Soft-MSD}(\text{KL}(P_{\text{img}} \parallel P_{\text{txt}}))$ & 55.6 & 61.1 & 56.7 & 52.6 & 51.9 & 0.267 & 0.190 \\
    $\text{Soft-MSD}(\text{Bi-KL}(\beta=0.5))$ & 56.4 & 62.6 & 57.8 & 53.3 & 52.4 & 0.292 & 0.209 \\
    $\text{Soft-MSD}\left(\text{KL}\left(P_{\text{txt}} \parallel P_{\text{img}}\right)\right)$ & \textbf{57.6} & \textbf{64.4} & \textbf{58.7} & \textbf{54.2} & \textbf{53.6} & \textbf{0.398} & \textbf{0.297} \\
    \thickhline
  \end{tabular}
\end{table*}

\textbf{Ambiguity-dependent effectiveness of Soft-MSD.}
We further analyze when Soft-MSD is beneficial by binning samples according to cosine margin \(\Delta \cos = \left| \cos(I,c^+) - \cos(I,c^-) \right|\). We observe that improvements concentrate in low-margin (high-ambiguity) cases, where global similarity is unreliable and local divergence becomes critical. This directly validates the uncertainty-aware design of Soft-MSD. Table~\ref{tab:cococf_bucket} shows that the improvements concentrate on the most ambiguous bins (Bin 0--1), supporting Soft-MSD's design as a confidence-aware fusion mechanism. 
On a 19,999-pair test set from COCO-CF Easy split, Soft-MSD improves over cosine by +4.70\% with 95\% bootstrap CI $[3.28, 6.12]$, and on a 19,993-pair test set from COCO-CF Hard split, Soft-MSD improves over cosine by +3.18\% with 95\% bootstrap CI $[2.04, 4.32]$.
To ensure statistical rigor, these CIs were computed via an \textbf{image-level} cluster bootstrap (resampling with replacement, $B=1000$), and a two-sided McNemar test with Edwards continuity correction further confirms that the improvements are statistically significant ($p<0.05$).

\begin{table*}[!t]
  \centering
  \renewcommand{\arraystretch}{1.25}
  \caption{Ambiguity-dependent effectiveness of Soft-MSD under cosine-margin partitioning (COCO-CF). Gains concentrate in low-confidence regimes, validating uncertainty-aware fusion.}
  \label{tab:cococf_bucket}
  \setlength{\tabcolsep}{8pt}
   \begin{tabular}{l|c|cc|cccc|c}
    \thickhline
    \textbf{Set} & \textbf{Bin} & \textbf{Mean $\Delta\cos$} & \textbf{Mean $u$} &
    \textbf{CLIPScore} & $\mathbf{KL}(P_{\text{img}}\|P_{\text{txt}})$ & \textbf{MSD} & \textbf{Soft-MSD} & \textbf{Gain} \\
    \hline
    
    \multirow{5}{*}{Easy-CF} 
      & 0 & 0.0017 & 0.996 & 52.28 & 60.78 & 61.31 & 61.31 & +9.03 \\
      & 1 & 0.0054 & 0.987 & 51.81 & 60.59 & 60.41 & 60.38 & +8.57 \\
      & 2 & 0.0095 & 0.976 & 58.58 & 63.74 & 63.55 & 63.39 & +4.81 \\
      & 3 & 0.0149 & 0.963 & 63.25 & 61.38 & 64.06 & 64.16 & +0.91 \\
      & 4 & 0.0266 & 0.934 & 70.03 & 59.50 & 70.31 & 70.19 & +0.16 \\
    \hline 
    \multirow{5}{*}{Hard-CF} 
      & 0 & 0.0017 & 0.996 & 53.33 & 59.24 & 59.24 & 59.27 & +5.94 \\
      & 1 & 0.0053 & 0.987 & 53.39 & 60.49 & 59.42 & 59.30 & +5.91 \\
      & 2 & 0.0092 & 0.977 & 54.38 & 58.44 & 57.57 & 57.79 & +3.41 \\
      & 3 & 0.0146 & 0.964 & 58.05 & 58.74 & 58.71 & 58.77 & +0.72 \\
      & 4 & 0.0264 & 0.934 & 61.86 & 57.96 & 61.77 & 61.80 & -0.06 \\
    
    \thickhline
  \end{tabular}
\end{table*}

\textbf{Generative captioner ranking via MSD.}
MSD can also serve as a discriminative signal for ranking caption generators without reference captions. Table~\ref{tab:gen_rank_ablation} shows the results, indicating consistent ordering across models, demonstrating that MSD captures global generation quality differences.

\begin{table}[!t]
  \centering
  \caption{Reference-free ranking of generative captioners using MSD.}
  \label{tab:gen_rank_ablation}
 \begin{tabular}{lc}
    \toprule
    \textbf{Captioner} & \textbf{Mean $\text{MSD}$} \\
    \midrule
    GPT-4o-mini \cite{hurst2024gpt}      & 0.2959 \\
    InternVL2-7B \cite{chen2024internvl}     & 0.2878 \\
    Qwen-VL-Chat \cite{wang2024qwen2}     & 0.2799 \\
    LLaVA-1.5-7B \cite{li2024llava}     & 0.2745 \\
    \bottomrule
  \end{tabular}
\end{table}

% \textbf{Systematic robustness across perturbations.}
% Figure~\ref{fig:ablation_models_subtask} presents the full SugarCrepe results across 54 settings. Across 54 compositional perturbation settings, Soft-MSD consistently outperforms cosine similarity, demonstrating robustness across model families and error types.

% \begin{figure*}[!t]
%   \centering
%   \includegraphics[width=\textwidth]{picture/table10.pdf}
%   \caption{Systematic evaluation of Soft-MSD across 54 compositional perturbation settings. Consistent gains demonstrate robustness across models and error types.}
%   \label{fig:ablation_models_subtask}
% \end{figure*}

\begin{figure}[!t]
  \centering
  \includegraphics[width=0.9\columnwidth]{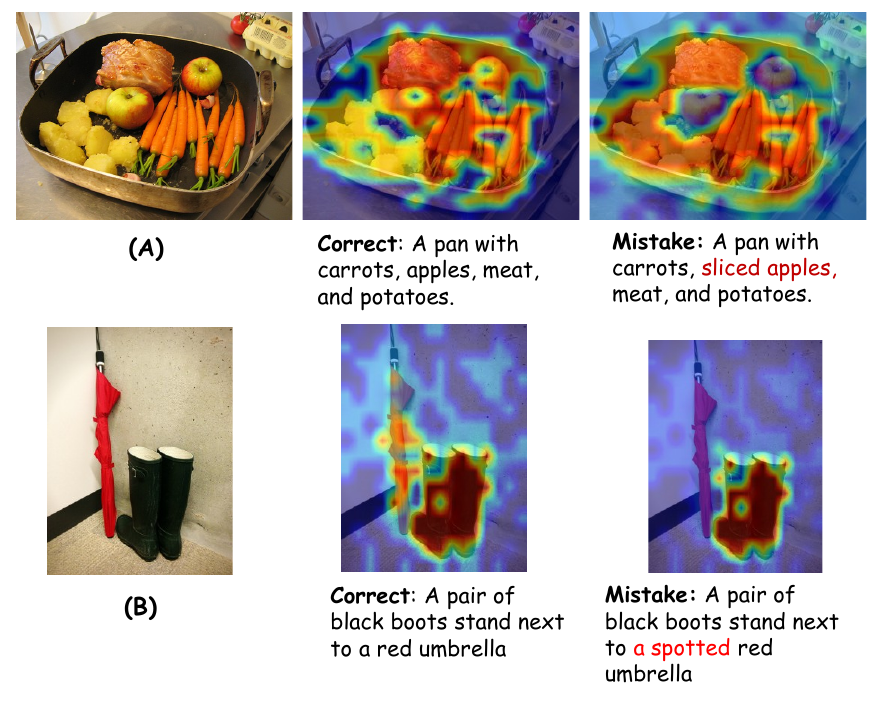}
  \vspace{-4mm}
  \caption{KL-decomposition-based interpretability of MSD showing semantic support and hallucination regions derived from bi-directional divergence.}
  \label{fig:vis1}
  \vspace{-4mm}
\end{figure}

\begin{figure*}[!t]
  \centering
  \includegraphics[width=0.9\textwidth,height=0.41\textheight,keepaspectratio]{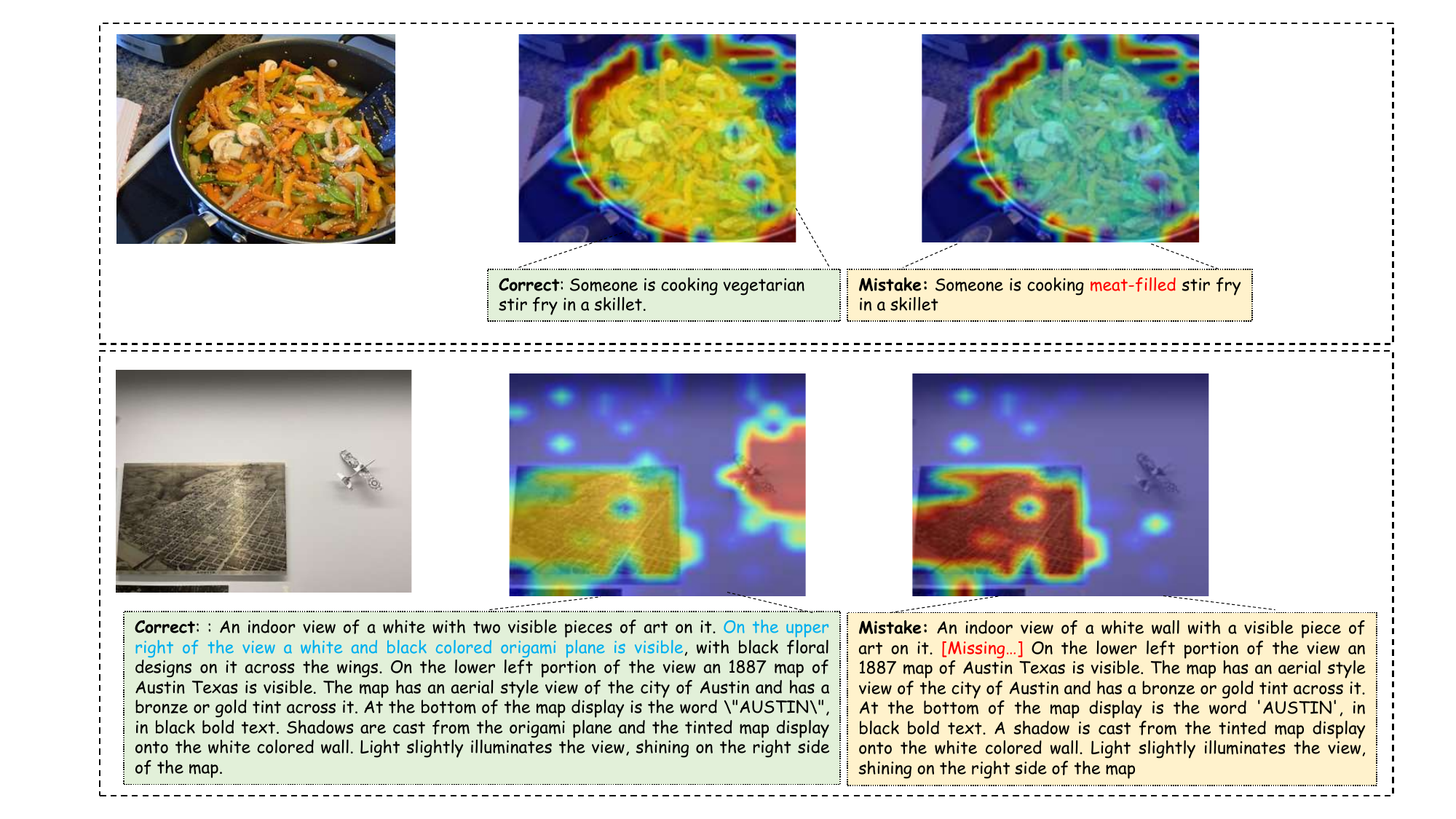}
  \vspace{-4mm}
  \caption{Qualitative examples. Failure modes including unsupported details and missing visual evidence, highlighted by token/region attribution.}
  \label{fig:qual_case_2}
  \vspace{-4mm}
\end{figure*}

\textbf{Interpretability via KL decomposition.}
Finally, MSD provides interpretable semantic attribution by decomposing bi-directional KL into patch-level contributions. Figure~\ref{fig:vis1} presents KL-decomposition-based heatmaps generated by Soft-MSD. Unlike attention-based visualizations, these heatmaps directly reflect divergence contributions. We decompose the Monte Carlo estimates of the two KL terms into local contributions, assign each patch/token a contribution score, and project the token-side support or penalty signal back to image space. Figure~\ref{fig:qual_case_2} shows failure cases, including unsupported details and missing visual evidence, with token/region attribution.

% Figure~\ref{fig:qual_case_1} highlights differences between correct captions and hard negatives in local grounding.
% Figure~\ref{fig:qual_case_2} shows failure cases, including unsupported details and missing visual evidence, with token/region attribution.
% Figure~\ref{fig:qual_case_3} focuses on fine-grained category confusions and omission-sensitive cases.
% Figure~\ref{fig:qual_case_4} demonstrates attribute errors and hallucinated objects in hard negatives.
% Figure~\ref{fig:qual_case_5} provides additional attribute-centric cases (e.g., color substitutions) where the hard negative introduces visually unsupported details.
% Collectively, these examples provide insight into the model’s strengths and typical error patterns.

To test whether the highlighted regions are score-relevant, we perform patch masking on SugarCrepe, results are shown in Table \ref{tab:heatmap_faithfulness}. For correct captions, we rank regions by the support map and report $\mathrm{BiKL}(\text{masked}) - \mathrm{BiKL}(\text{original})$; for incorrect captions, we rank regions by the penalty map and report $\mathrm{BiKL}(\text{original}) - \mathrm{BiKL}(\text{masked})$. We can see that removing high-contribution regions significantly affects scores, while low-contribution regions have negligible impact.

\begin{table}[!t]
  \centering
  \renewcommand{\arraystretch}{1.15}
  \caption{Faithfulness validation of KL-decomposed attribution via patch masking on SugarCrepe.}
  \label{tab:heatmap_faithfulness}
  \resizebox{0.48\textwidth}{!}{
  \begin{tabular}{l|l|ccc}
    \thickhline
    \textbf{Caption type} & \textbf{Map} & \textbf{Top-10\% mask} & \textbf{Random-10\% mask} & \textbf{Bottom-10\% mask} \\
    \hline
    Correct & Support & \textbf{0.1159} & 0.0531 & $-0.0209$ \\
    Incorrect & Penalty & \textbf{0.0350} & $-0.0477$ & $-0.1052$ \\
    \thickhline
  \end{tabular}}
\end{table}

% \begin{figure*}[!t]
%   \centering
%   \includegraphics[width=\textwidth,height=0.41\textheight,keepaspectratio]{picture/appendix1.pdf}
%   \caption{Qualitative examples 1. Additional cases illustrating local grounding differences between the correct caption and a hard-negative caption.}
%   \label{fig:qual_case_1}
% \end{figure*}

% \begin{figure*}[!t]
%   \centering
%   \includegraphics[width=\textwidth,height=0.41\textheight,keepaspectratio]{picture/appendix3.pdf}
%   \caption{Qualitative examples 3. Fine-grained category confusions and omission-sensitive cases.}
%   \label{fig:qual_case_3}
% \end{figure*}

% \begin{figure*}[!t]
%   \centering
%   \includegraphics[width=\textwidth,height=0.41\textheight,keepaspectratio]{picture/appendix4.pdf}
%   \caption{Qualitative examples 4. Additional cases covering attribute errors and hallucinated objects in hard negatives.}
%   \label{fig:qual_case_4}
% \end{figure*}

% \begin{figure*}[!t]
%   \centering
%   \includegraphics[width=\textwidth,height=0.41\textheight,keepaspectratio]{picture/appendix5.pdf}
%   \caption{Qualitative examples 5. Additional attribute-level hard negatives (e.g., color substitutions) and the corresponding attribution patterns.}
%   \label{fig:qual_case_5}
% \end{figure*}

\subsection{Discussion and Limitations}

The core motivation of MSD-Score is to address a specific gap in reference-free evaluation: the lack of a \emph{decomposable and controllable} signal for fine-grained semantic verification. While prior metrics and MLLM-as-a-Judge approaches both capture aspects of caption quality, they exhibit complementary limitations. Global embedding-based metrics (e.g., CLIPScore) collapse image--text correspondence into a single similarity score, making them insensitive to localized semantic errors such as omissions or hallucinations. On the other hand, judge-based methods provide strong holistic judgments, but their decision process is implicit, non-deterministic, and difficult to attribute or control. MSD-Score is designed as a middle ground between these paradigms. By explicitly modeling local distributions and measuring bi-directional divergence, it provides a \emph{structured and interpretable verification signal} that complements holistic similarity. In particular, it enables: (i) explicit decomposition into coverage (image$\rightarrow$text) and support (text$\rightarrow$image), (ii) deterministic and reproducible scoring without reliance on generative inference, and (iii) controllable behavior under different regimes (e.g., caption length or ambiguity). We therefore view MSD-Score as a \emph{complementary evaluation layer} rather than a replacement for judge-based methods.

MSD provides interpretability through KL-based decomposition into patch/token contributions, offering a mechanistic view of where mismatches arise. However, we emphasize that this form of interpretability is \emph{diagnostic rather than semantic}: it explains \emph{why the score changes} in terms of local evidence alignment, but does not fully capture higher-level reasoning or linguistic coherence. As such, the provided heatmaps should be interpreted as a tool for analyzing grounding and hallucination, rather than a complete explanation of caption quality.

% Like all embedding-based metrics, MSD-Score inherits limitations from the underlying vision--language encoder. Under significant domain shift (e.g., medical or remote sensing imagery), performance depends on the quality of the pretrained representations. However, the proposed framework is modular: the alignment component can be adapted once per domain, while the scoring formulation (vMF + bi-directional KL) remains unchanged. This makes MSD-Score a \emph{domain-adaptable evaluation layer}, rather than a dataset-specific metric.

More broadly, the proposed multi-scale distributional formulation is not limited to caption evaluation. The ability to explicitly model local alignment and measure bidirectional support suggests potential applications in other generative evaluation settings, such as text-to-image or multimodal reasoning, where fine-grained consistency between modalities is critical. At the same time, we acknowledge that MSD-Score represents a step toward structured evaluation rather than a complete solution, and further integration with higher-level reasoning signals remains an important direction for future work.

\section{Conclusion}

We present MSD-Score, a reference-free image caption metric that formulates image--text alignment as multi-scale distributional matching. By modeling patch- and token-level embeddings as hyperspherical mixtures, MSD-Score preserves fine-grained semantics lost in holistic metrics. Weighted bi-directional divergence captures complementary failure modes, and uncertainty-aware fusion integrates local verification with global relevance. Experiments across benchmarks show strong correlation with human judgments and improved detection of fine-grained errors. The distributional formulation provides transparent, deterministic, and decomposable signals that complement both holistic similarity metrics and judge-based evaluators. MSD-Score advances faithful and robust evaluation for vision--language generation models.

% \section*{Acknowledgments}

%%%%%%%%%%%%%%%%%%%%%%%%%%%%%%%%%%%%%%%%%%%%%%%%%%%%%%%%%%%%

% \section*{Impact Statement}
% This paper presents work whose goal is to advance the field of Machine Learning. There are many potential societal consequences of our work, none which we feel must be specifically highlighted here.

% \nocite{langley00} % removed for IEEEtran compile: entry absent from example_paper.bib

\bibliographystyle{IEEEtran}
\bibliography{example_paper}

\end{document}